\documentclass[conference,compsoc]{IEEEtran}

\renewcommand\IEEEkeywordsname{Keywords}

\ifCLASSINFOpdf
  \usepackage[pdftex]{graphicx}
  \graphicspath{ {pictures/}{../pdf/}{../jpeg/} }
  \DeclareGraphicsExtensions{.pdf,.jpeg,.png,.PNG}
\else
\fi
\ifCLASSOPTIONcompsoc
 \usepackage[caption=false,font=normalsize,labelfont=sf,textfont=sf]{subfig}
\else
 \usepackage[caption=false,font=footnotesize]{subfig}
\fi
\usepackage{flushend}
\usepackage{tabularx,booktabs,textcomp}
\newcolumntype{C}{>{\centering\arraybackslash}X} 
\newcolumntype{L}{>{\raggedright\arraybackslash}X} 
\usepackage[noadjust]{cite}

\usepackage{wasysym}
\usepackage{lipsum} 
\usepackage{amssymb}

\usepackage{color}
\PassOptionsToPackage{hyphens}{url}
\usepackage{hyperref}
\begin{document}
\title{Brain Tumor Classifiers Under Attack: Robustness of ResNet Variants Against Transferable FGSM and PGD Attacks}


\makeatletter
\newcommand{\linebreakand}{%
  \end{@IEEEauthorhalign}
  \hfill\mbox{}\par
  \mbox{}\hfill\begin{@IEEEauthorhalign}
}
\makeatother

\author{
  \IEEEauthorblockN{Ryan Deem}
  \IEEEauthorblockA{\textit{Department of Computer Science} \\
    \textit{Kennesaw State University}\\
    Marietta, GA, USA \\
    rdeem@students.kennesaw.edu}
  \and
  \IEEEauthorblockN{Garrett Goodman}
  \IEEEauthorblockA{\textit{Computer Science and Software Engineering} \\
    \textit{Miami University}\\
    Oxford, OH, USA \\
    goodmag@miamioh.edu}
  \linebreakand 
  \IEEEauthorblockN{Waqas Majeed}
  \IEEEauthorblockA{\textit{Department of Computer Science} \\
    \textit{Kennesaw State University}\\
    Marietta, GA, USA \\
    wmajeed@kennesaw.edu}
  \and
  \IEEEauthorblockN{Md Abdullah Al Hafiz Khan}
  \IEEEauthorblockA{\textit{Department of Computer Science} \\
    \textit{Kennesaw State University}\\
    Marietta, GA, USA \\
    mkhan74@kennesaw.edu}
  \and
  \IEEEauthorblockN{Michail S. Alexiou}
  \IEEEauthorblockA{\textit{Department of Computer Science} \\
    \textit{Kennesaw State University}\\
    Marietta, GA, USA \\
    malexiou@kennesaw.edu}
}



\maketitle

\IEEEpubidadjcol

\begin{abstract}
Adversarial robustness in deep learning models for brain tumor classification remains an underexplored yet critical challenge, particularly for clinical deployment scenarios involving MRI data. In this work, we investigate the susceptibility and resilience of several ResNet-based architectures, referred to as BrainNet, BrainNeXt and DilationNet, against gradient-based adversarial attacks, namely FGSM and PGD. These models, based on ResNet, ResNeXt, and dilated ResNet variants respectively, are evaluated across three preprocessing configurations (i) full-sized augmented, (ii) shrunk augmented and (iii) shrunk non-augmented MRI datasets. Our experiments reveal that BrainNeXt models exhibit the highest robustness to black-box attacks, likely due to their increased cardinality, though they produce weaker transferable adversarial samples. In contrast, BrainNet and Dilation models are more vulnerable to attacks from each other, especially under PGD with higher iteration steps and $\alpha$ values. Notably, shrunk and non-augmented data significantly reduce model resilience, even when the untampered test accuracy remains high, highlighting a key trade-off between input resolution and adversarial vulnerability. These results underscore the importance of jointly evaluating classification performance and adversarial robustness for reliable real-world deployment in brain MRI analysis.
\end{abstract}

\begin{IEEEkeywords}
Adversarial Attacks; Adversarial Transferability; Adversarial Robustness; Brain Tumor Classification; ResNet; ResNeXt; Dilated Convolutional Neural Networks;
\end{IEEEkeywords}




%



\section{Introduction}
\label{s:intro}

Brain tumors pose a critical health challenge globally, affecting individuals across all demographics and age groups. In the United States alone, nearly 94,390 new cases of primary brain and other central nervous system (CNS) tumors were diagnosed in 2023, with approximately 18,990 of them being fatal\cite{cbrs-nih-article}. Gliomas account for about 27\% of all brain tumors and 80\% of malignant ones, while meningiomas and pituitary tumors represent 38\% and 16\% respectively\cite{cancer-facts-figs-data-website}. These statistics underscore the need for accurate and reliable diagnostic tools to assist early detection and treatment planning. In recent years, CNNs and other deep learning architectures have shown significant promise in automating brain tumor detection from MRI scans\cite{medical_ex_ictai} among other applications\cite{face_ex_ictai, autonomous_ex_ictai}. However, despite their widespread adoption, these models remain vulnerable to adversarial attacks, which are subtlly crafted perturbations that can lead to incorrect predictions with high confidence \cite{szegedy2014}. This vulnerability is particularly concerning in medical settings, where model decisions may have life-altering consequences. Among the most widely studied adversarial techniques are the Fast Gradient Sign Method (FGSM) \cite{goodfellow2015} and its iterative counterpart, Projected Gradient Descent (PGD) \cite{madry2018towards}, which leverage gradients to construct small perturbations that can fool even highly accurate models. As the adoption of deep learning grows in clinical decision-making pipelines, ensuring the reliability and security of such models, especially under black-box settings, becomes increasingly critical. This raises a pressing question:  \textit{Do faster, more efficient medical imaging models come at the cost of robustness when facing white box and black-box adversarial threats?}

\begin{figure}[ht]
   \centering
    \includegraphics[height=0.25\textheight, width=0.45\textwidth]{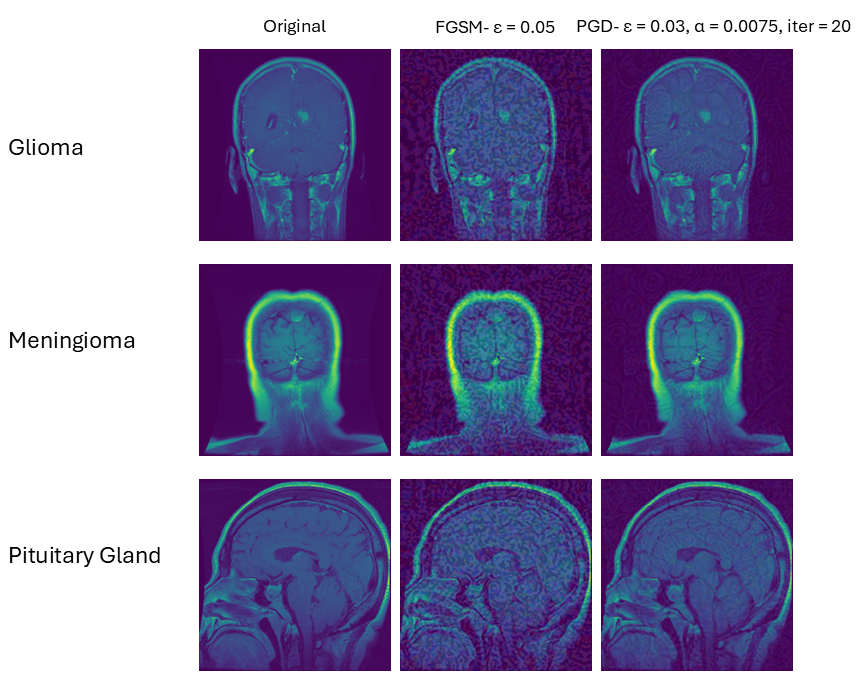}
    \caption{Examples of original and adversarial samples generated using FGSM and PGD, shown across different parameter settings (i.e., $\epsilon$, $\alpha$, and number of iterations).}
    \label{Dataset 1's images}
\end{figure}

\begin{figure}[ht]
   \centering
    \includegraphics[height=0.25\textheight, width=0.45\textwidth]{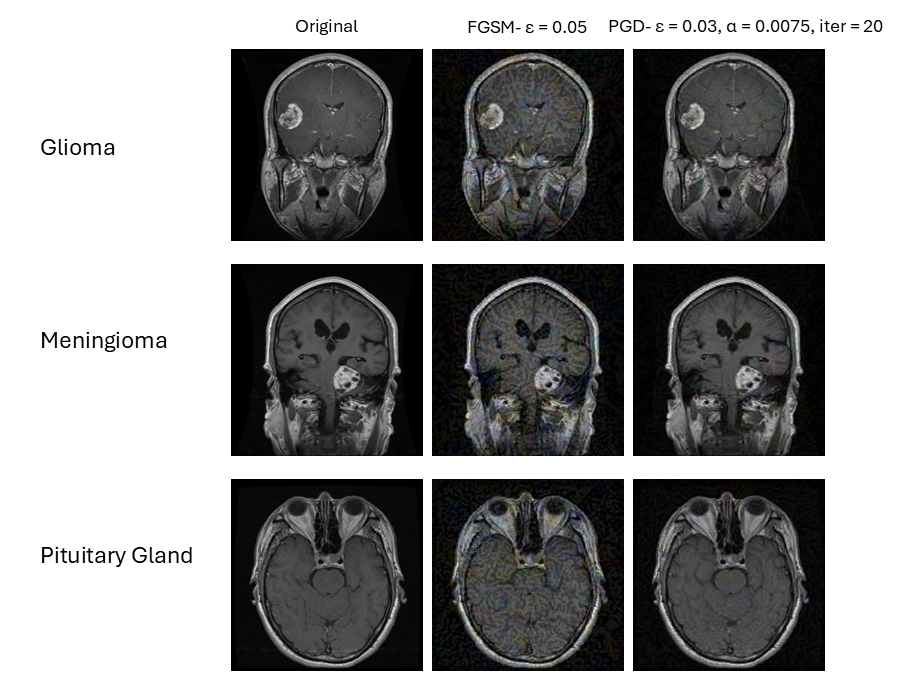}
    \caption{Additional examples of original and adversarial samples generated using FGSM and PGD, shown across different parameter settings (i.e., $\epsilon$, $\alpha$, and number of iterations).}
    \label{Dataset 2's images}
\end{figure}

Kotia et al. \cite{white_box_medical_imaging} examine the vulnerability of CNN-based brain tumor classifiers to white-box attacks like Noise Injection, FGSM, and VAT, emphasizing the risk to clinical reliability. They show that adversarial training improves robustness. Expanding on this, Joel et al. \cite{Joel2022_AdversarialOncology} test VGG16-based models on CT, MRI, and mammogram data under FGSM, BIM, and PGD, revealing similar weaknesses. Yinusa et al. \cite{yinusa2025multi} also target VGG16-based classifiers with FGSM and PGD, proposing a defense that combines ensemble adversarial training and feature squeezing (e.g., bit-depth reduction, Gaussian blurring), significantly restoring accuracy under attack. Collectively, these studies highlight a core issue: classifiers inherit vulnerabilities from backbone architectures like VGG16, MobileNet, and EfficientNet, making them especially prone to transferable attacks, which is an important concern for safe clinical deployment.

Wei et al. \cite{csecs_2023} identify limitations in gradient-based attacks like FGSM and I-FGSM, and propose ABINM, a novel method that avoids the gradient sign function and leverages averaged gradient accumulation across iterations to boost both strength and transferability. Similarly, Khalid et al. \cite{ichora_2025} evaluate the robustness of a custom CNN and ResNet-50 against various perturbations, including FGSM, Gaussian noise, and boundary noise. To defend against these, they introduce Adversarial Attack-Driven Data Augmentation, integrating perturbed samples into training to improve resilience. Both works highlight how even structurally similar CNNs can be vulnerable to transferable attacks, yet the robustness of newer, faster CNN variants in black-box settings remains underexplored.

In this paper, we shift our focus from accuracy and computational efficiency \cite{iisa_ryan, ictai_alex} to robustness under adversarial conditions, investigating how gradient-based attacks affect CNN-based architectures for MRI-based brain tumor classification. While deep learning models like ResNet and ResNeXt have demonstrated high accuracy in classifying glioma, meningioma, and pituitary tumors, their vulnerability to adversarial perturbations, especially in clinical imaging scenarios, remains a critical concern. To this end, we conduct a systematic evaluation of three model families: BrainNet (ResNet-based), BrainNeXt (ResNeXt-based), and Dilation-based (ResNet with dilated convolutions), analyzing their performance under both white-box and black-box attack settings using FGS and PGD attacks. We also evaluate their resilience against attacks generated on DenseNet for further insights. Our experiments span different image configurations, including full-sized images, resized (shrunk) images and augmented vs non-augmented versions, to help investigate further the impact of input quality on adversarial transferability. Across these settings, we examine not only attack effectiveness, but also the transferability of adversarial examples between architectures. All developed architectures are evaluated using the brain tumor MRI dataset provided in \cite{dataset1}. Representative image samples from this dataset are shown in Fig.~\ref{Dataset 1's images} and Fig.~\ref{Dataset 2's images} respectively , including both original inputs and their corresponding adversarial variants generated via FGSM and PGD attacks across each tumor class. The aim of this work is to uncover how different architectural choices and data preprocessing strategies influence model robustness and attack transferability, offering practical insights for building safer systems in medical imaging applications.


\section{Literature Review}
\label{s:relawork}

\subsection{Brain Tumor Classification}
Nazir et al. \cite{nazir-iccs} propose an efficient brain tumor classification pipeline that combines wavelet decomposition with DCT-based feature extraction and a neural network classifier, achieving strong performance with low computational cost. Hafeez et al. \cite{hafeez-siu} design a custom 15-layer CNN for classifying meningioma, glioma, and pituitary tumors using contrast-enhanced T1-weighted MRIs, outperforming several standard CNNs, though their model is tailored to a specific dataset. While such handcrafted or block-based pipelines are effective, they often lack adaptability across imaging modalities or tumor types. In contrast, Mehemud et al. \cite{mehemud-iccit} employ transfer learning with DenseNet121 and EfficientNetB0 to classify MRIs into four categories, including non-tumor, outperforming traditional classifiers and VGG16. Li et al. \cite{li-itoec} enhance EfficientNet-B0 with an MSCAM attention module and MBConv tweaks to improve multi-scale feature learning. Govind et al. \cite{govind} also apply transfer learning, fine-tuning a CNN for four tumor classes—including a custom GMNP class, achieving strong generalization with limited labeled data.

\subsection{Adversarial Transferability \& Defenses}

Several recent studies have explored strategies to improve adversarial transferability in black-box settings. Wu et al. \cite{wu_2020} introduced Attention-Guided Transfer Attacks (ATA), leveraging gradient-based attention to focus on shared internal features, reducing source model overfitting. They later proposed ATTA \cite{wu_2021}, a CNN-based transformation network that simulates distortions to improve adversarial durability. Xiong et al. \cite{xiong_2022} developed SVRE, an ensemble approach using stochastic optimization to minimize gradient variance and boost transfer strength. Yang et al. \cite{yang_2023} presented Dark Surrogate Models (DSMs), trained with soft labels and data mixing to enhance generalization. Yin et al. \cite{yin_2024} proposed a meta-learning framework using a Meta Conditional Generator (MCG) to adapt to new attack scenarios from prior experience. On the defense side, Sun et al. \cite{bo_sun2019} introduced a transformation layer that filters high-frequency noise while preserving key features. Kafali et al. \cite{kafali2021} proposed efficient adversarial training by skipping attack iterations, reducing overhead without sacrificing robustness. Ouyang et al. \cite{ouyang2022} introduced BT-Dropout, randomly dropping internal layers during training to promote stable feature learning. Zhang et al. \cite{brianzhang2022} proposed norm shaping to reinforce decision boundaries by maintaining mostly clean inputs per batch. Lastly, Hsiung et al. \cite{hsiung2023} presented a composite adversarial training method that dynamically mixes attack types (e.g., noise, rotations, brightness) using a randomized generator during training.

\section{Method Overview}
\label{s:implementation}
We designed seven CNN models based on the ResNet family to analyze MRI data, including BrainNet (a modified ResNet-101) and its variants BrainNeXt-50, -101, and -152, which follow the ResNeXt architecture. Some models incorporated dilated convolutions in later layers to improve multi-scale feature extraction. All models used early stopping after six epochs without validation loss improvement.
\subsection{Model Architectures and Variations}
\subsubsection{Base Model}
The base BrainNet model is built on the ResNet-101 architecture pre-trained on ImageNet. To enhance feature extraction, we added two additional residual blocks. During training, the initial layers were frozen for 20 epochs with early stopping enabled after 6 stagnant epochs. Subsequently, layers from index 327 onward were unfrozen, and the model was retrained for another 20 epochs under the same early stopping criteria. ResNet was chosen as the backbone due to its demonstrated strength in preserving feature information across layers. We selected ResNet-101 over deeper variants like ResNet-152 to minimize the risk of overfitting on our dataset.


\subsubsection{BrainNeXt Models}
In contrast to BrainNet, the BrainNeXt models are based on the ResNeXt architecture, which emphasizes width over depth for more efficient feature extraction. This is achieved through cardinality—each residual block consists of multiple parallel convolutional paths (using smaller filters), all processing the same input. The outputs are then aggregated and merged with the original input to form the final block output. Compared to the standard ResNet design, this structure enables BrainNeXt to capture richer feature representations with fewer parameters. To thoroughly assess its effectiveness and avoid overfitting, we evaluated all three standard ResNeXt configurations. 


\subsubsection{Dilation CNN Models}
The final architectural variation we explored in the original BrainNet model involved introducing dilation into the two residual blocks that were initially included. In standard CNNs, convolutional layers use filters that slide over input images to detect spatial hierarchies, beginning with simple local features like edges and textures. As the network deepens, these features are progressively combined to recognize more complex patterns. Pooling layers then reduce the spatial dimensions of these representations, preserving the most salient information while discarding redundancy. A typical CNN layer produces a feature map $F$ by convolving the input $X$ with a filter $W$ and applying a non-linear activation function $\sigma$, as shown in Eq. \ref{eq:eq1}:
\begin{equation}
F = \sigma(W * X + b)\label{eq:eq1}
\end{equation}
where $*$ denotes the convolution operation, $b$ is the bias term, and $\sigma$ is typically a ReLU activation.\vspace{1\baselineskip}

Dilated CNNs extend standard CNNs by introducing a dilation factor $d$, which enlarges the receptive field and allows the network to capture more context without increasing the number of parameters. In a dilated convolution, the filter is applied over input elements spaced apart by $d$, as shown in Eq. \ref{eq:eq2}:
\begin{equation}
F[i] = \sum_{k=1}^{K} W[k] \cdot X[i + d \cdot k]\label{eq:eq2}
\end{equation}
where $d$ is the dilation rate and $K$ is the kernel size. This approach is especially useful for capturing multi-scale context in tasks requiring dense predictions. The configurations and runtimes for each Dilated CNN variation are as follows: with dilation rates of 2, 3, and 4, all models had $42,626,560$ parameters, took 15 minutes to train, and had an inference time of 3 seconds.

\subsection{Threat Model and Attack Formulation}
This study assumes a black-box adversarial setting, where the attacker lacks access to the target model’s architecture, weights, or gradients. Instead, adversarial examples are crafted using a separate surrogate model and transferred to the target, leveraging the well-known phenomenon of adversarial transferability \cite{szegedy2014, papernot2016transferabilitymachinelearningphenomena}. This setup reflects real-world scenarios where model internals are typically hidden. We use two common gradient-based attacks—FGSM and PGD—to generate adversarial inputs from each model, which are then evaluated on all others to assess cross-model robustness. Experiments are conducted on three MRI dataset variations: full-sized augmented, shrunk augmented, and shrunk non-augmented. Each attack is tested with $\epsilon$ values of 0.02, 0.03, 0.04, and 0.05, where $\epsilon$ controls perturbation magnitude. For normalized images in the [0,1] range, $\epsilon = 0.04$ corresponds to ±10 pixel intensity change on a 0–255 scale.\\

\noindent\textbf{Fast Gradient Sign Method (FGSM)} creates adversarial examples by perturbing the input in the direction of the loss gradient’s sign, scaled by $\epsilon$ \cite{goodfellow2015}. The perturbed input $x_{\text{adv}}$ is computed as:
\begin{equation}
x_{\text{adv}} = x + \epsilon \cdot \text{sign}(\nabla_x J(\theta, x, y))\label{eq:eq5}
\end{equation}

Here, $x$ is the original input, $y$ the true label, $\theta$ the model parameters, and $J$ the loss function.

\noindent\textbf{Projected Gradient Descent (PGD)} improves upon FGSM by applying the perturbation iteratively, starting from a randomly perturbed input \cite{madry2018towards}. This helps escape local minima and improves attack success. PGD is defined as:
\begin{align}
x_0^{\text{adv}} &= x + \mathcal{U}(-\epsilon, \epsilon), \label{eq:eq7a} \\
x_{t+1}^{\text{adv}} &= \Pi_{B_\epsilon(x)}\left(x_t^{\text{adv}} + 
\alpha \cdot \text{sign}\left(\nabla_x J(\theta, x_t^{\text{adv}}, y)\right)\right) \label{eq:eq7b}
\end{align}

Here, $\mathcal{U}$ denotes uniform random noise and $\Pi_{B_\epsilon(x)}(\cdot)$ projects the input back into the $\epsilon$-ball around $x$.

\subsection{Hyperparameters and Training Configuration}
We used the following hyperparameters consistently across all models: a learning rate of 0.0001, a batch size of 10, and a total of 150 training epochs. The Adam optimizer was used in conjunction with the Sparse Categorical Cross-Entropy loss function. Let $y \in \mathbb{R}^C$ be the one-hot encoded true label vector and $\hat{y} \in \mathbb{R}^C$ the predicted probability distribution from the model’s softmax output. The loss function $J(\theta, x, y)$ is given by Eq. \ref{eq:eq4}:
\begin{equation}
J(\theta, x, y) = -\sum_{i=1}^{C} y_i \log(\hat{y}_i)\label{eq:eq4}
\end{equation}
where $x$ is the input sample, $y_i$ is the true label indicator (1 if class $i$ is correct, 0 otherwise), $\hat{y}_i = \text{softmax}(f_\theta(x))_i$ is the predicted probability for class $i$, $f_\theta(x)$ is the model's raw output (logits), $\theta$ represents the model parameters and $C$ is the number of classes. Early stopping was enabled with a patience of 6 epochs, monitoring validation loss to prevent overfitting. The only deviation from these hyperparameters was during the fine-tuning phase of the BrainNet models, where the learning rate was reduced by a factor of 10.

\subsection{Hardware and Software Details}
The following hardware and software configurations were employed to train and evaluate the models:
\textbf{Hardware:} GPU: NVIDIA GeForce RTX 4090, CPU: AMD Ryzen 9 7950X3D, RAM: 64 GB DDR5, Storage: 4 TB SSD.
\textbf{Software:} Operating System: Windows 11 Pro (host), Ubuntu 22.04.03 LTS (via WSL2), Framework: TensorFlow 2.14, Programming Language: Python 3.10.12, CUDA Version: 11.5, Other Libraries: NumPy, Matplotlib, scikit-learn.

\section{Evaluation and Results}
\subsection{Preprocessing and Variations}
During the evaluation stage, we used a publicly available MRI dataset consisting of 4,023 images spanning three common brain tumor types: glioma, meningioma, and pituitary tumors \cite{dataset1}. All images in the dataset are 24-bit colored MRI scans. While the original dataset also includes a fourth class representing benign (non-tumor) cases, we excluded this class from training. As a result, the model may exhibit a slight bias toward the glioma class due to its higher frequency in the remaining dataset. To address potential class imbalance and ensure robust evaluation, we split the dataset into 60\% training, 10\% validation, 20\% testing, and 10\% attack subsets. We applied several preprocessing pipelines to investigate their impact on model robustness. Models with suffix -3 or v3 were trained on augmented data, which included random flips and rotations. Models labeled with -5 or v5 were trained on the non-augmented version of the same dataset. Additionally, we explored the effect of input resolution. The original MRI images, sized at 512×512×3, were referred to as the full-sized versions. These were compared to shrunk versions resized to 160×160×3. Accordingly, we use the terms full-sized and shrunk to distinguish between models trained on these different input dimensions.

\begin{equation}
\alpha=\epsilon/4\label{eq:alpha_eq_1}
\end{equation}
\begin{equation}
\alpha=\epsilon/iterations\label{eq:alpha_eq_2}
\end{equation}
For the PGD attack variations, we employed two different formulas to compute the step size parameter, $\alpha$. The first approach, shown in Equation~\ref{eq:alpha_eq_1}, provides a more aggressive step size, particularly at the iteration counts explored in our experiments. The second approach, presented in Equation~\ref{eq:alpha_eq_2}, defines a variable $\alpha$ that adapts based on the number of iterations. This formulation is more suitable for higher iteration counts, as it scales the perturbation step to maintain stability and effectiveness over longer attack sequences.

\subsection{Results on FGSM Attacks}
FGSM's strength as an attack lies in the effectiveness of its gradient-based formulation combined with fine-tuning of the $\epsilon$ parameter, which controls the intensity of the perturbation. To investigate how different attack strengths influence adversarial transferability while preserving imperceptibility, we perform a series of experiments across a range of $\epsilon$ values. These tests aim to evaluate the robustness of three key representative models: BrainNeXt-152 (based on ResNeXt), BrainNet (based on ResNet), and Dilation3 (a ResNet variant with dilated CNN layers of $dilation\;rate=3$). More importantly, adversarial samples are generated from these models and their architectural variations (seven in total), thus, emulating both white-box and black-box evaluation settings. To further examine how image quality impacts transferability effectiveness, we conduct experiments across three dataset variants: (i) full-sized augmented MRI images, (ii) shrunk augmented MRI images for efficiency and (iii) shrunk non-augmented MRI images.

\begin{figure}[ht]
   \centering
    \includegraphics[height=0.18\textheight, width=0.45\textwidth]{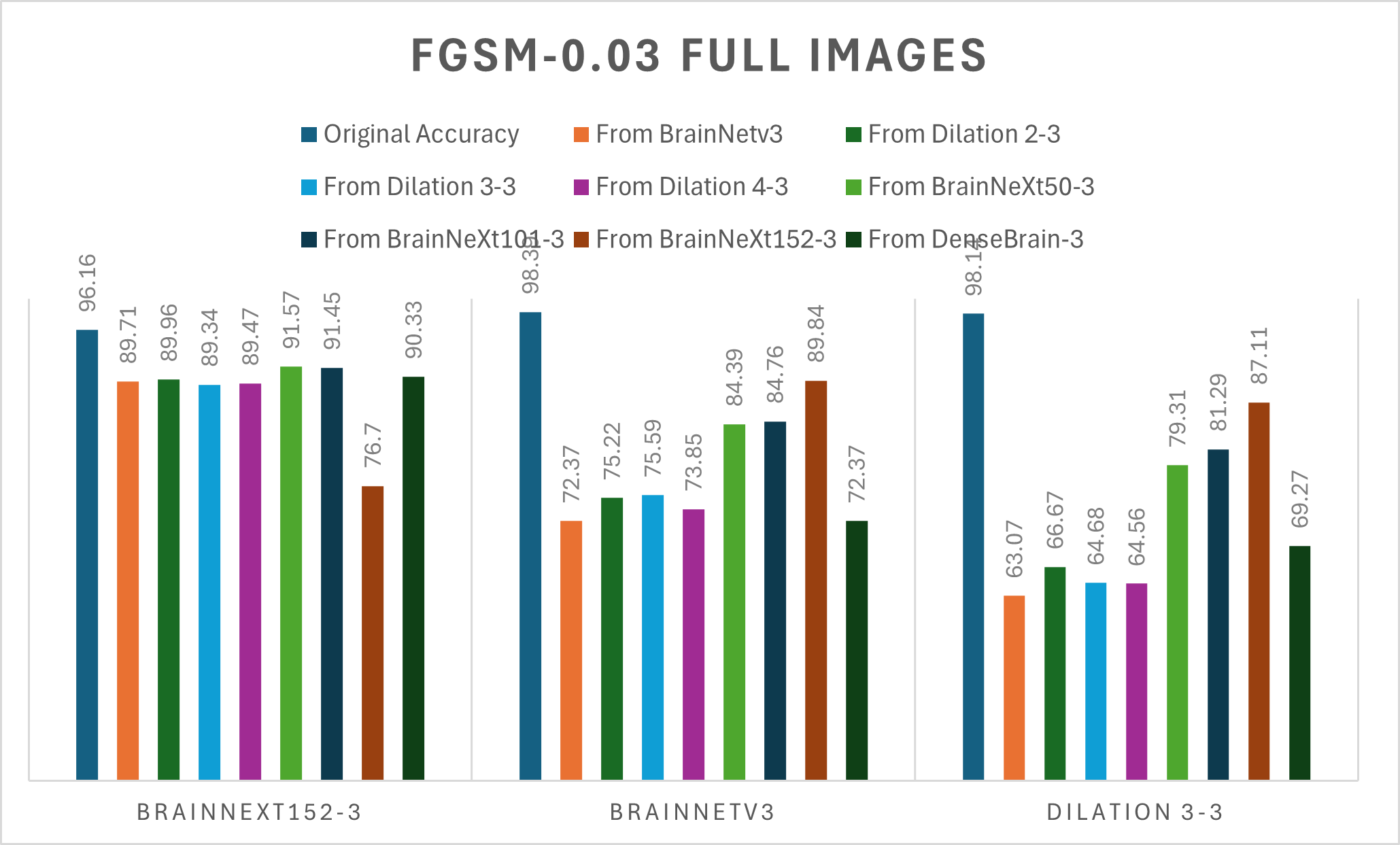}
    \caption{Accuracy of the BrainNeXt152-3, BrainNetv3, and Dilation3-3 models under FGSM white-box and black-box attacks using the full-sized and augmented MRI dataset, evaluated at $\epsilon = 0.03$.}
    \label{FGSM-0.03-1}
\end{figure}
\begin{figure}[ht]
   \centering
    \includegraphics[height=0.18\textheight, width=0.45\textwidth]{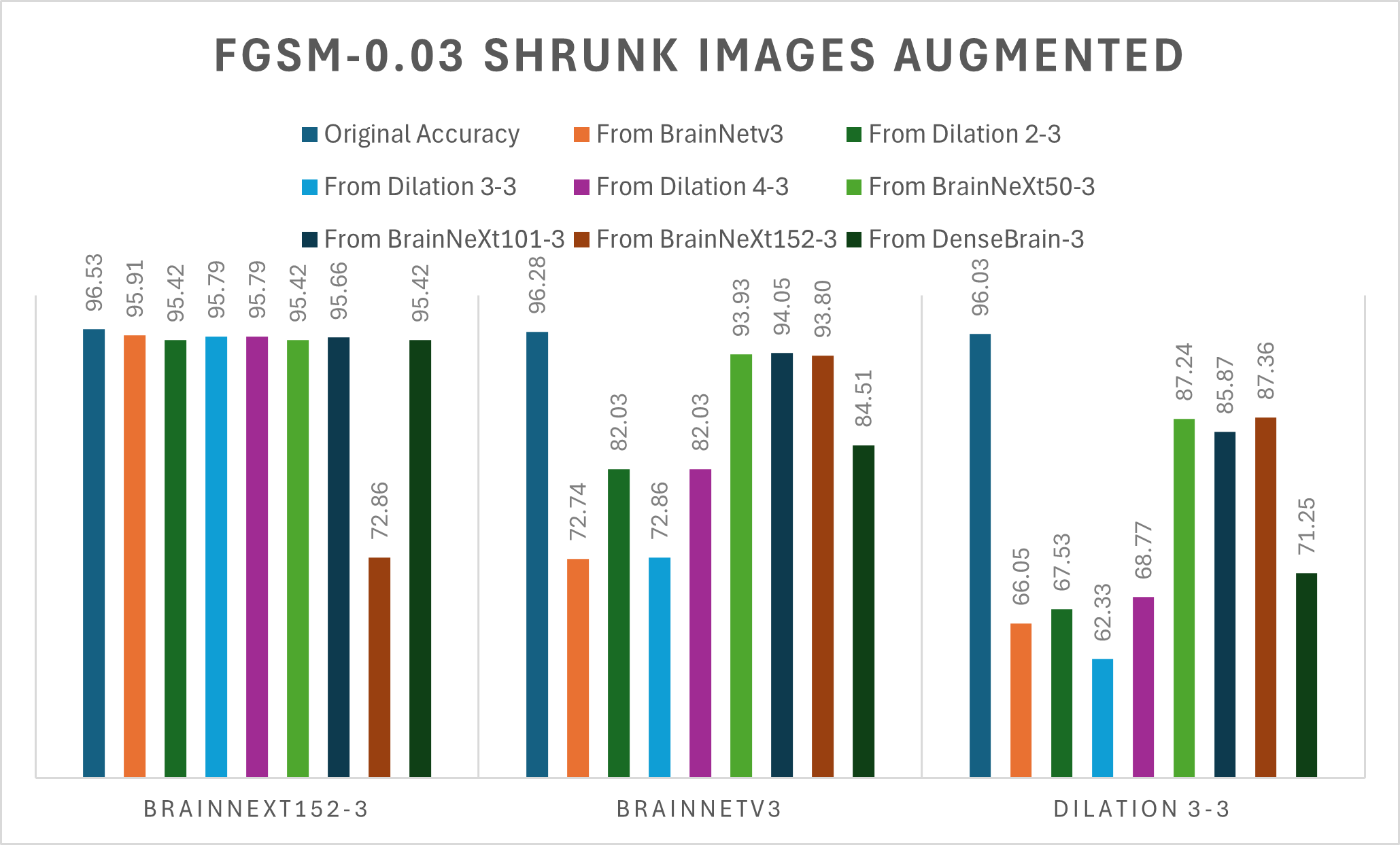}
    \caption{Accuracy of the BrainNeXt152-3, BrainNetv3, and Dilation3-3 models under FGSM white-box and black-box attacks using the shrunk and augmented MRI dataset, evaluated at $\epsilon = 0.03$.}
    \label{FGSM-0.03-2}
\end{figure}
\begin{figure}[ht]
   \centering
    \includegraphics[height=0.18\textheight, width=0.45\textwidth]{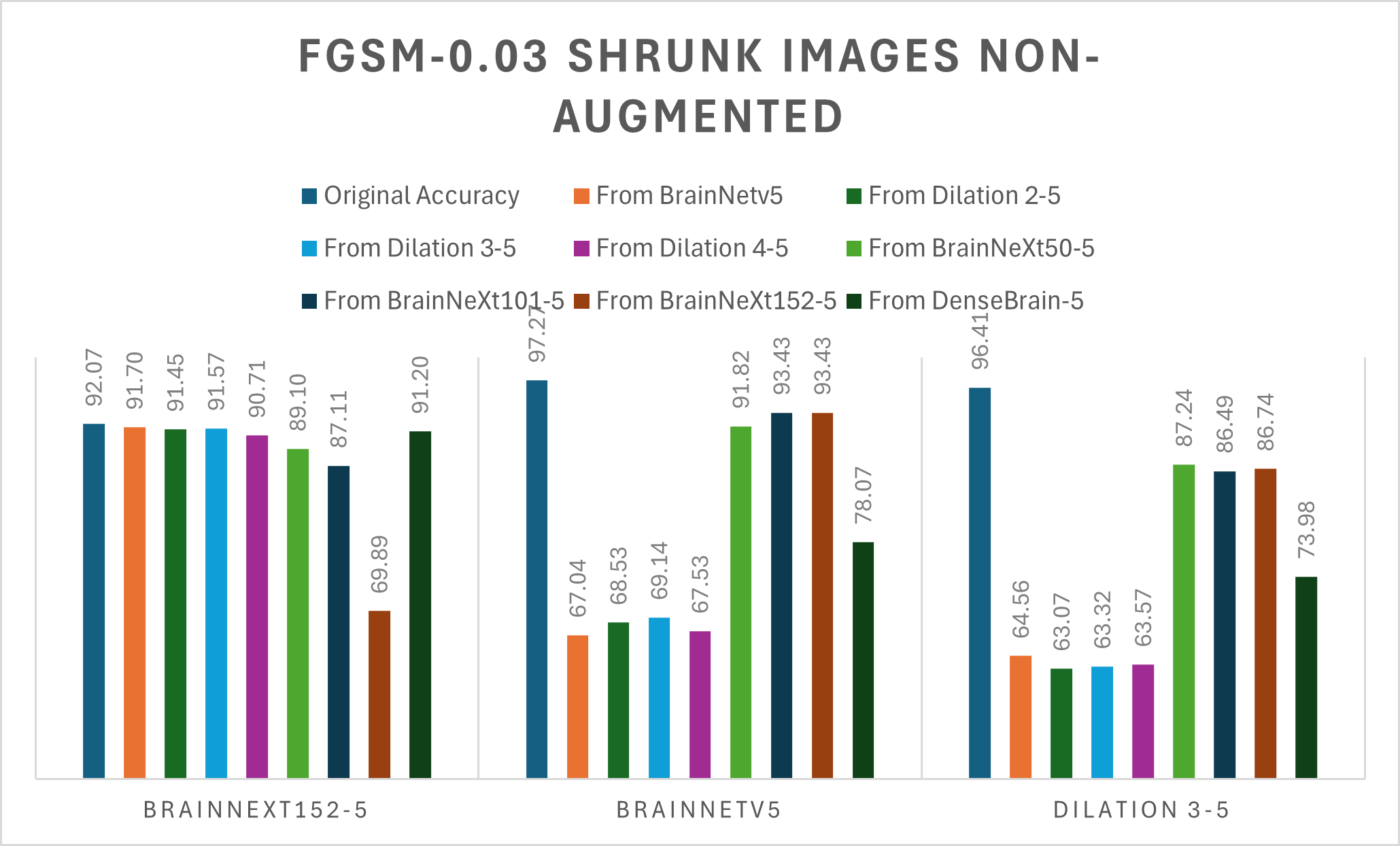}
    \caption{Accuracy of the BrainNeXt152-5, BrainNetv5, and Dilation3-5 models under FGSM white-box and black-box attacks using the shrunk and non-augmented MRI dataset, evaluated at $\epsilon = 0.03$.}
    \label{FGSM-0.03-3}
\end{figure}


Fig.~\ref{FGSM-0.03-1}, Fig.~\ref{FGSM-0.03-2} and Fig.~\ref{FGSM-0.03-3} present the accuracy of the BrainNeXt152-5, BrainNet, and Dilation3 models under FGSM white-box and black-box attacks, evaluated at $\epsilon = 0.03$, across three dataset variants: (i) full-sized augmented, (ii) shrunk augmented, and (iii) shrunk non-augmented MRI images, respectively. Across all experiments, the BrainNeXt152 architecture consistently demonstrated high resilience. In particular, on the full-sized augmented dataset, its accuracy dropped by approximately 22\% under a white-box attack, yet it still outperformed all other models in the same setting. Similar robustness was observed in the shrunk augmented and shrunk non-augmented settings, where BrainNeXt152 maintained accuracy above 87\% under black-box attacks, with the largest drop occurring when attacked by BrainNeXt101 on the shrunk non-augmented data.

In contrast, BrainNetv3 and Dilation3-3 exhibited a different trend. Both models were relatively robust against adversarial examples generated by BrainNeXt variants, showing accuracy drops of 2–7\% (BrainNet) and 6–11\% (Dilation3) across all datasets. However, both models were significantly more vulnerable to attacks crafted by each other. These vulnerabilities became especially apparent in the shrunk augmented and shrunk non-augmented datasets. More specifically, Dilation3's accuracy dropped by 23–35\% when attacked using adversarial examples generated by other Dilation variants (white-box with dilation rate 3 or black-box with dilation rates 2 and 4), and by approximately 32\% under black-box attacks from BrainNet, all within the shrunk augmented setting. Performance degradation was even more pronounced in the shrunk non-augmented case. Dilation3's accuracy fell below 70\% against both white-box and black-box attacks, with the worst-case accuracy reaching approximately 63\% under a white-box attack. BrainNet experienced a similar trend, although its performance decline was less severe compared to Dilation3. These results highlight the increased vulnerability of both models under reduced image quality and data augmentation, especially when facing attacks originating from more structurally similar models.

\begin{figure}[ht]
   \centering
    \includegraphics[height=0.18\textheight, width=0.45\textwidth]{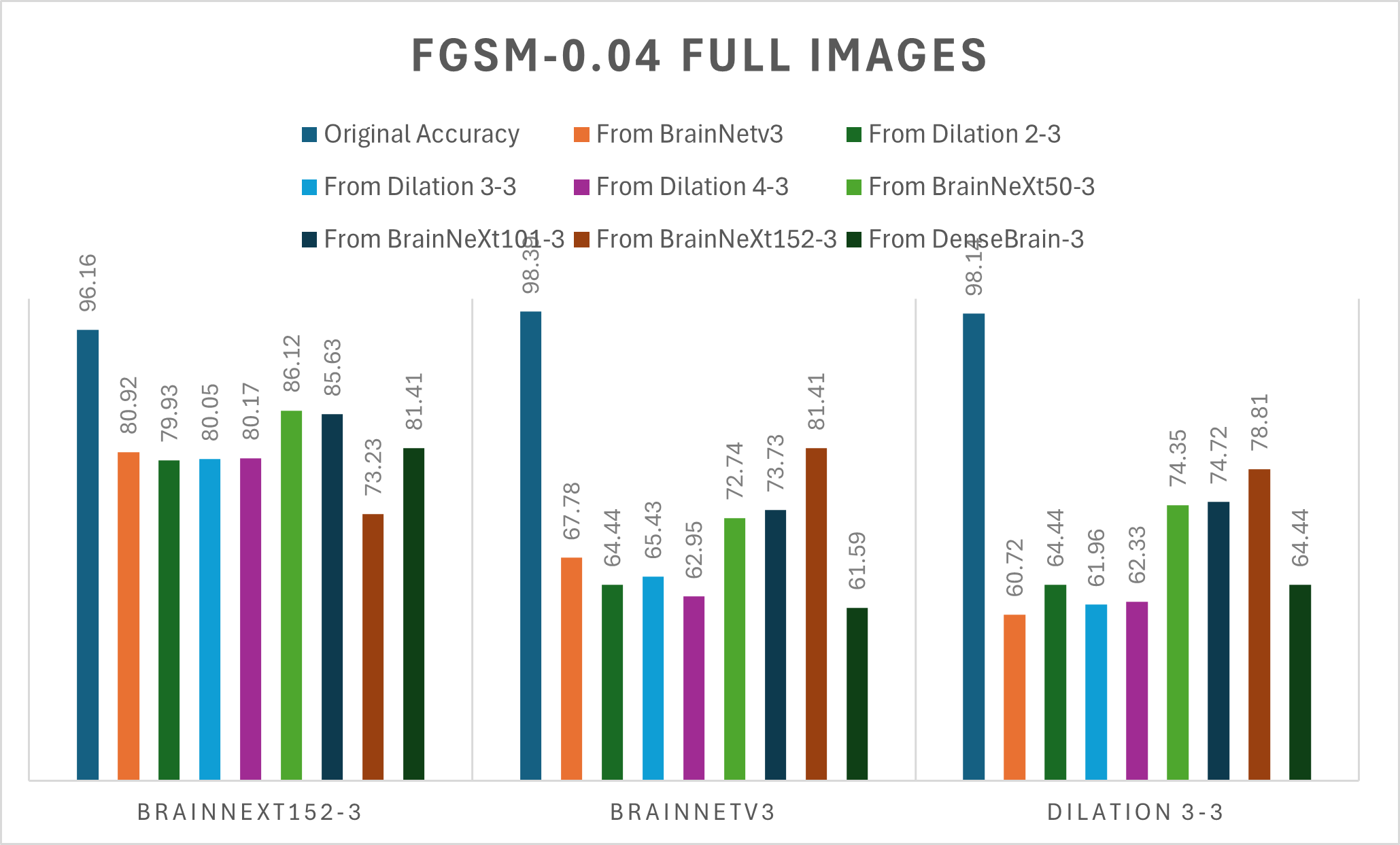}
    \caption{Accuracy of the BrainNeXt152-3, BrainNetv3, and Dilation3-3 models under FGSM white-box and black-box attacks using the full-sized and augmented MRI dataset, evaluated at $\epsilon = 0.04$.}
    \label{FGSM-0.04-1}
\end{figure}
\begin{figure}[ht]
   \centering
    \includegraphics[height=0.18\textheight, width=0.45\textwidth]{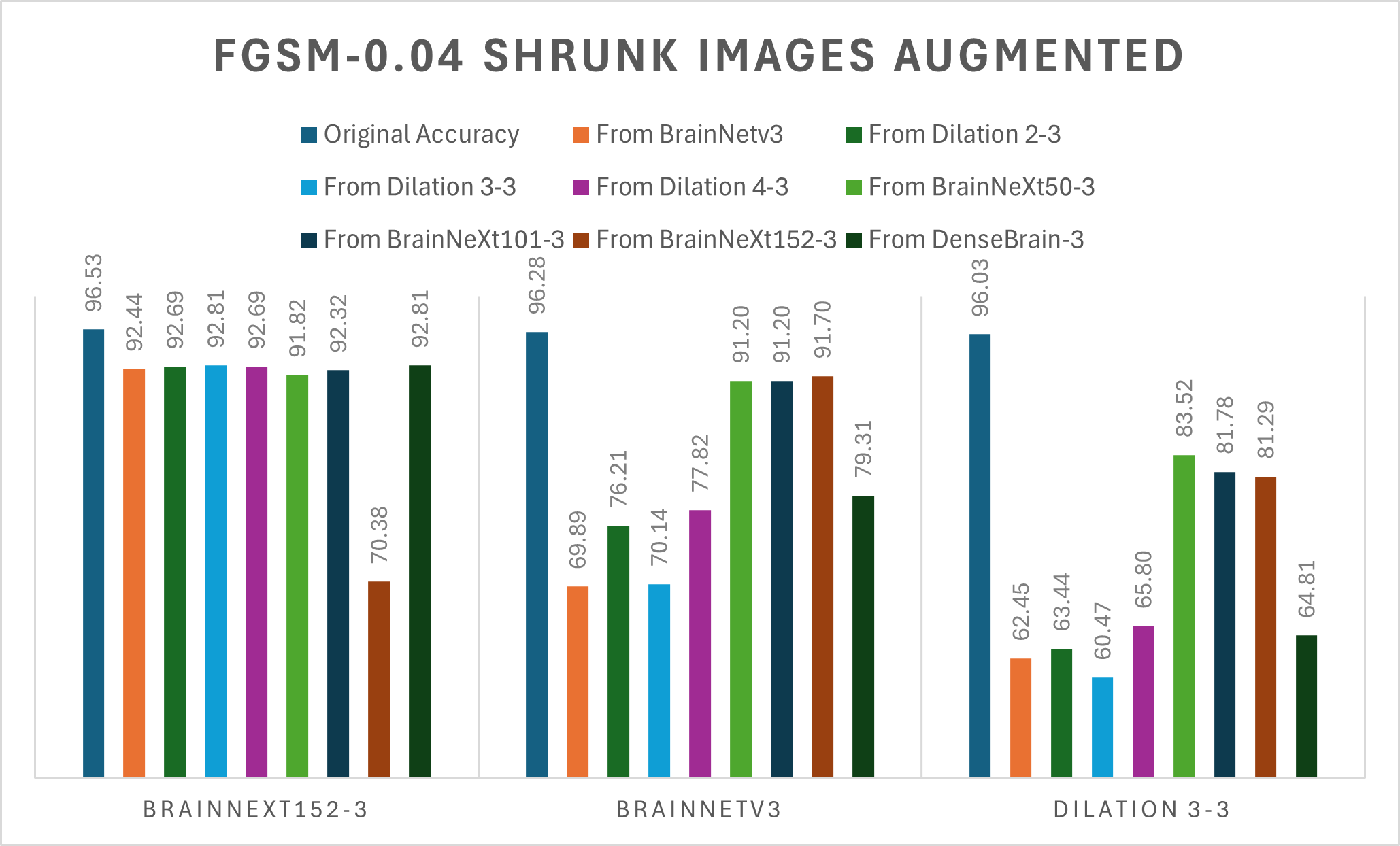}
    \caption{Accuracy of the BrainNeXt152-3, BrainNetv3, and Dilation3-3 models under FGSM white-box and black-box attacks using the shrunk and augmented MRI dataset, evaluated at $\epsilon = 0.04$.}
    \label{FGSM-0.04-2}
\end{figure}
\begin{figure}[ht]
   \centering
    \includegraphics[height=0.18\textheight, width=0.45\textwidth]{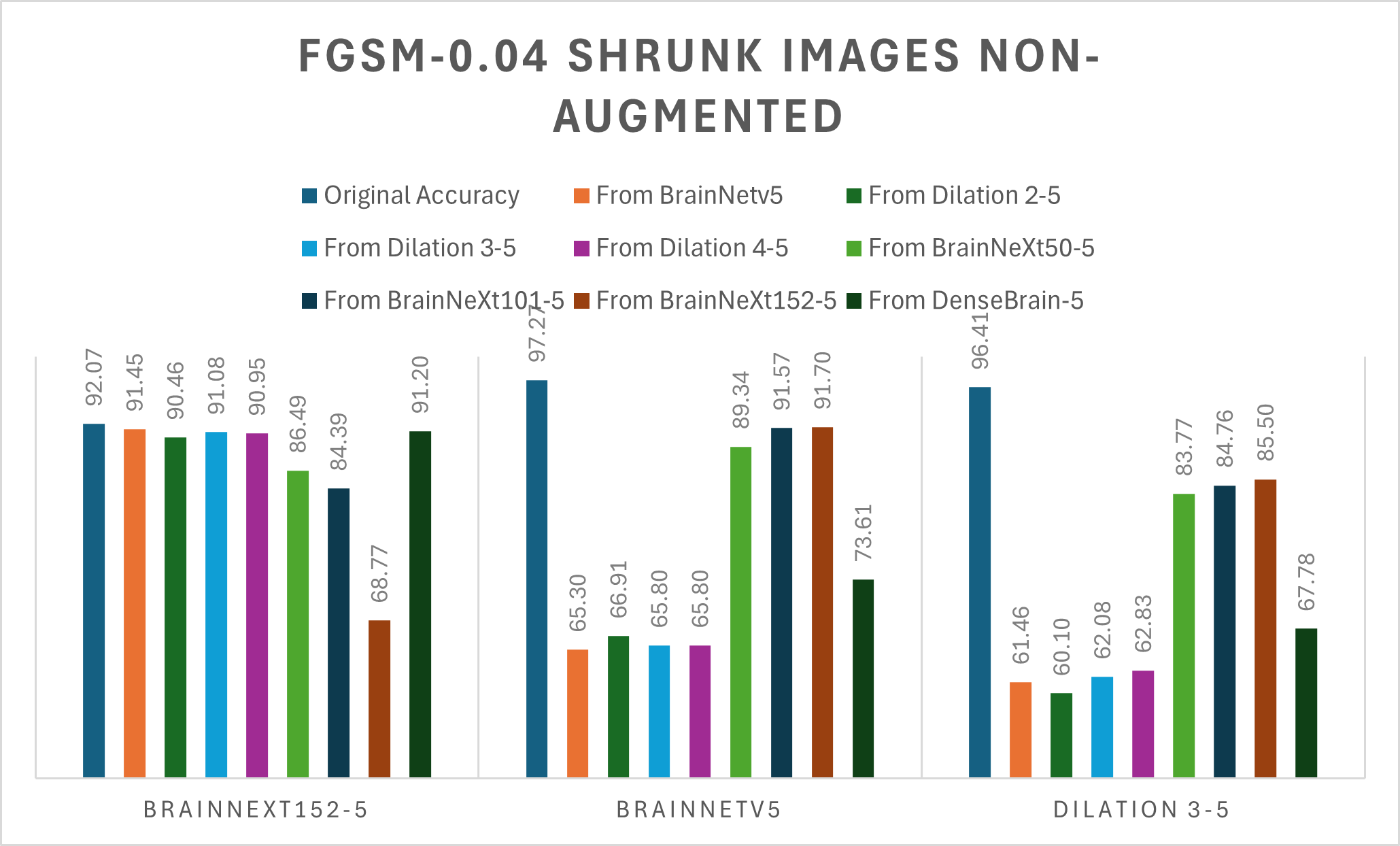}
    \caption{Accuracy of the BrainNeXt152-5, BrainNetv5, and Dilation3-5 models under FGSM white-box and black-box attacks using the shrunk and non-augmented MRI dataset, evaluated at $\epsilon = 0.04$.}
    \label{FGSM-0.04-3}
\end{figure}

Finally, Fig.~\ref{FGSM-0.04-1}, Fig.~\ref{FGSM-0.04-2}, and Fig.~\ref{FGSM-0.04-3} show the accuracy of the BrainNeXt152-5, BrainNet, and Dilation3 models under FGSM white-box and black-box attacks at $\epsilon = 0.04$, across three dataset variants: (i) full-sized augmented, (ii) shrunk augmented, and (iii) shrunk non-augmented MRI images. Similarly, Fig.~\ref{FGSM-0.05-1}, Fig.~\ref{FGSM-0.05-2}, and Fig.~\ref{FGSM-0.05-3} present results for the same models and dataset variations under $\epsilon = 0.05$. Across both sets of experiments, we observe consistent patterns, with BrainNeXt152 continuing to demonstrate the strongest resilience to black-box attacks but also exhibiting the weakest transferability, producing less effective adversarial samples when targeting BrainNet and Dilation-based models. In contrast, Dilation3 and BrainNet show sharper drops in accuracy under black-box attacks crafted from other Dilation variants (Dilation2 and Dilation4) and BrainNet itself—effects that are further amplified by the increased $\epsilon$ value. At $\epsilon = 0.05$, the lowest black-box accuracy for Dilation3 is recorded at 59.98\% from attacks generated by Dilation2, while BrainNet drops to 64.19\% under attacks from Dilation3, both using the shrunk non-augmented dataset. These results represent accuracy declines of approximately 35–37\% respectively for these two models. These results further reinforce the initial observation that the shrunk non-augmented dataset leads to overall reduced model robustness, across all three architectures in black-box attack settings for MRI tumor classification. Lastly, it must be noted that across all FGSM experiments, BrainNeXt consistently demonstrated resilience against DenseNet-based attacks, with only about a 0.1\% accuracy drop. In contrast, both Dilation3 and BrainNet exhibited significant performance degradation, particularly when evaluated on the shrunk augmented dataset.

\begin{figure}[ht]
   \centering
    \includegraphics[height=0.18\textheight, width=0.45\textwidth]{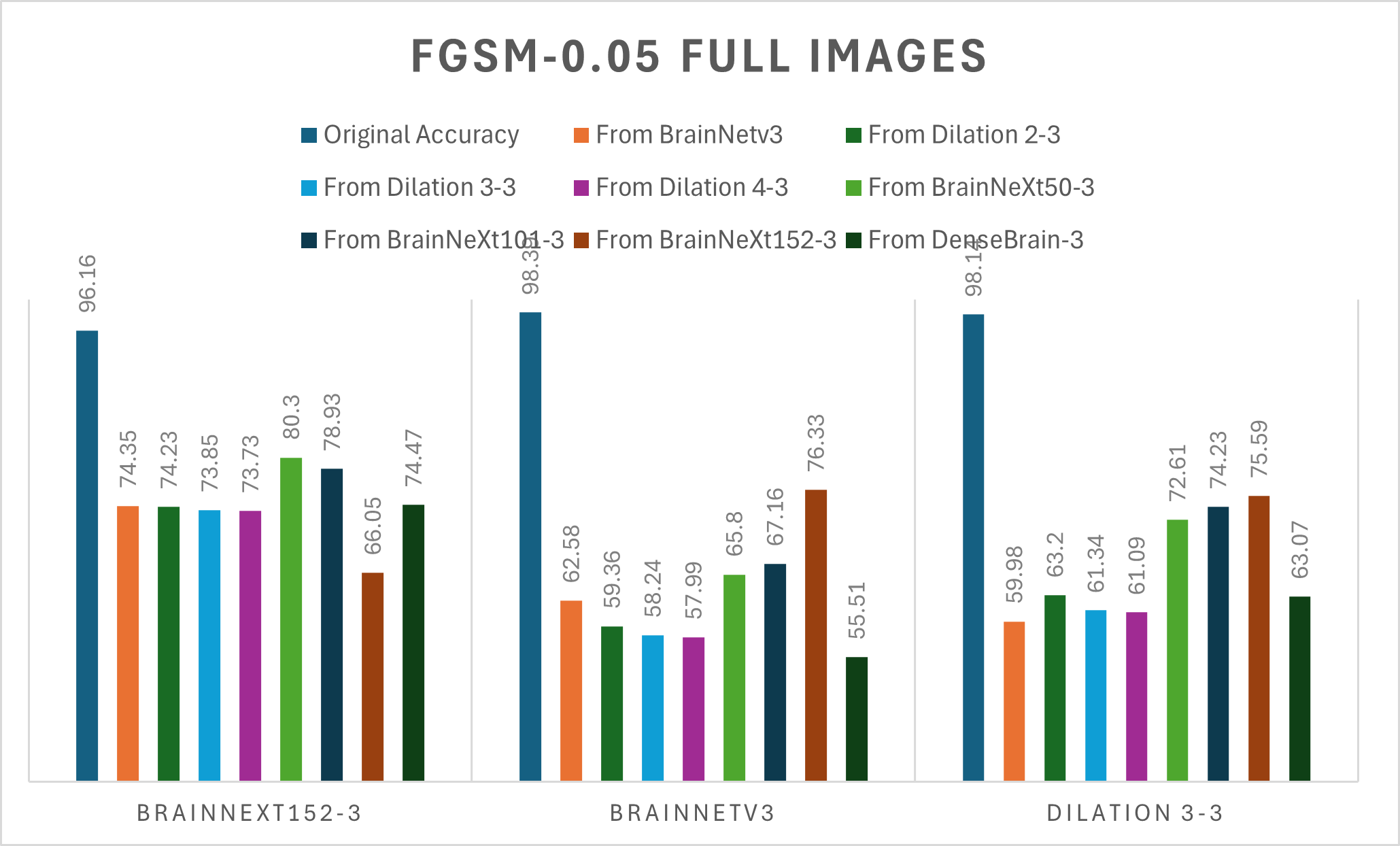}
    \caption{Accuracy of the BrainNeXt152-3, BrainNetv3, and Dilation3-3 models under FGSM white-box and black-box attacks using the full-sized and augmented MRI dataset, evaluated at $\epsilon = 0.05$.}
    \label{FGSM-0.05-1}
\end{figure}
\begin{figure}[ht]
   \centering
    \includegraphics[height=0.18\textheight, width=0.45\textwidth]{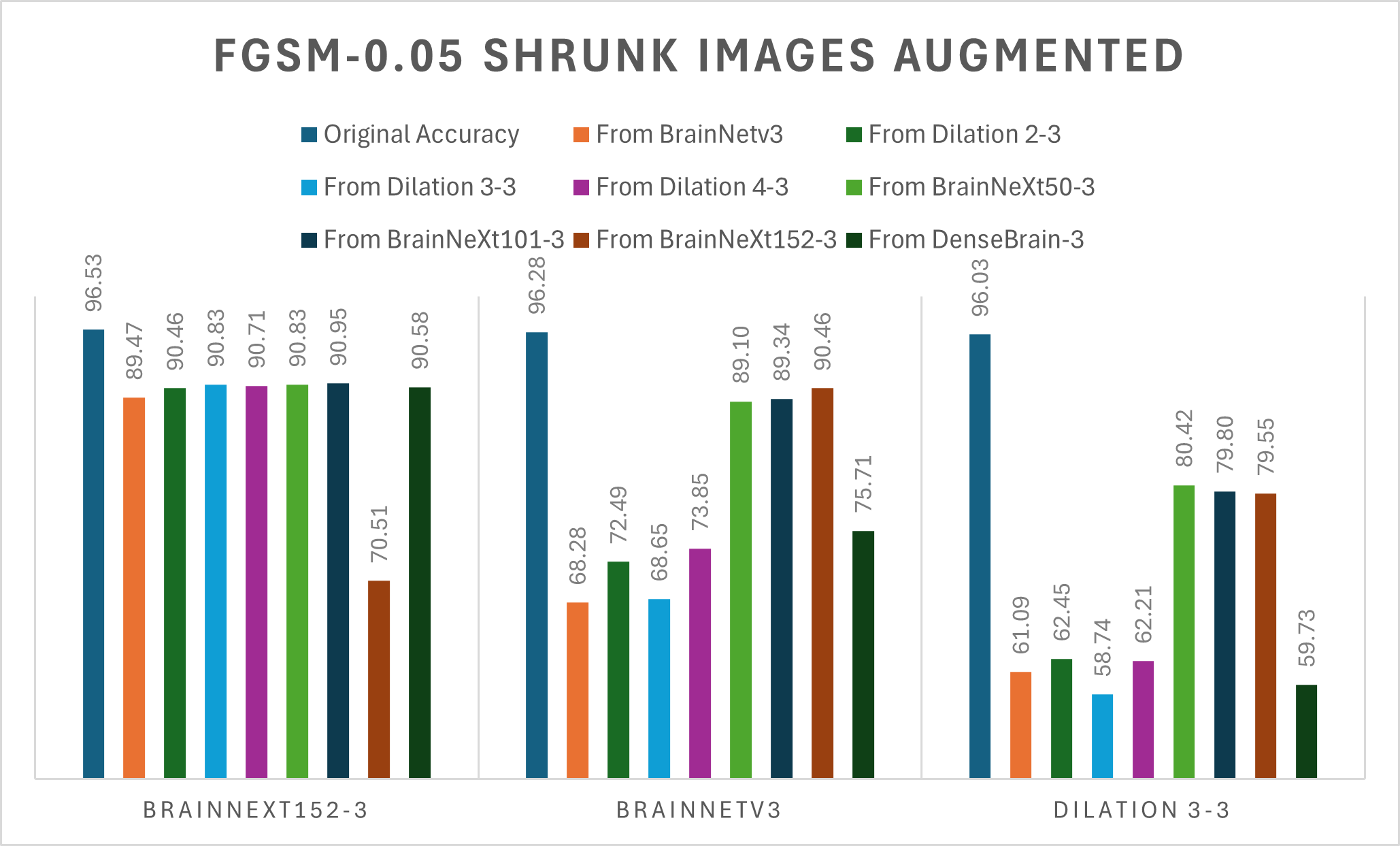}
    \caption{Accuracy of the BrainNeXt152-3, BrainNetv3, and Dilation3-3 models under FGSM white-box and black-box attacks using the shrunk and augmented MRI dataset, evaluated at $\epsilon = 0.05$.}
    \label{FGSM-0.05-2}
\end{figure}
\begin{figure}[ht]
   \centering
    \includegraphics[height=0.18\textheight, width=0.45\textwidth]{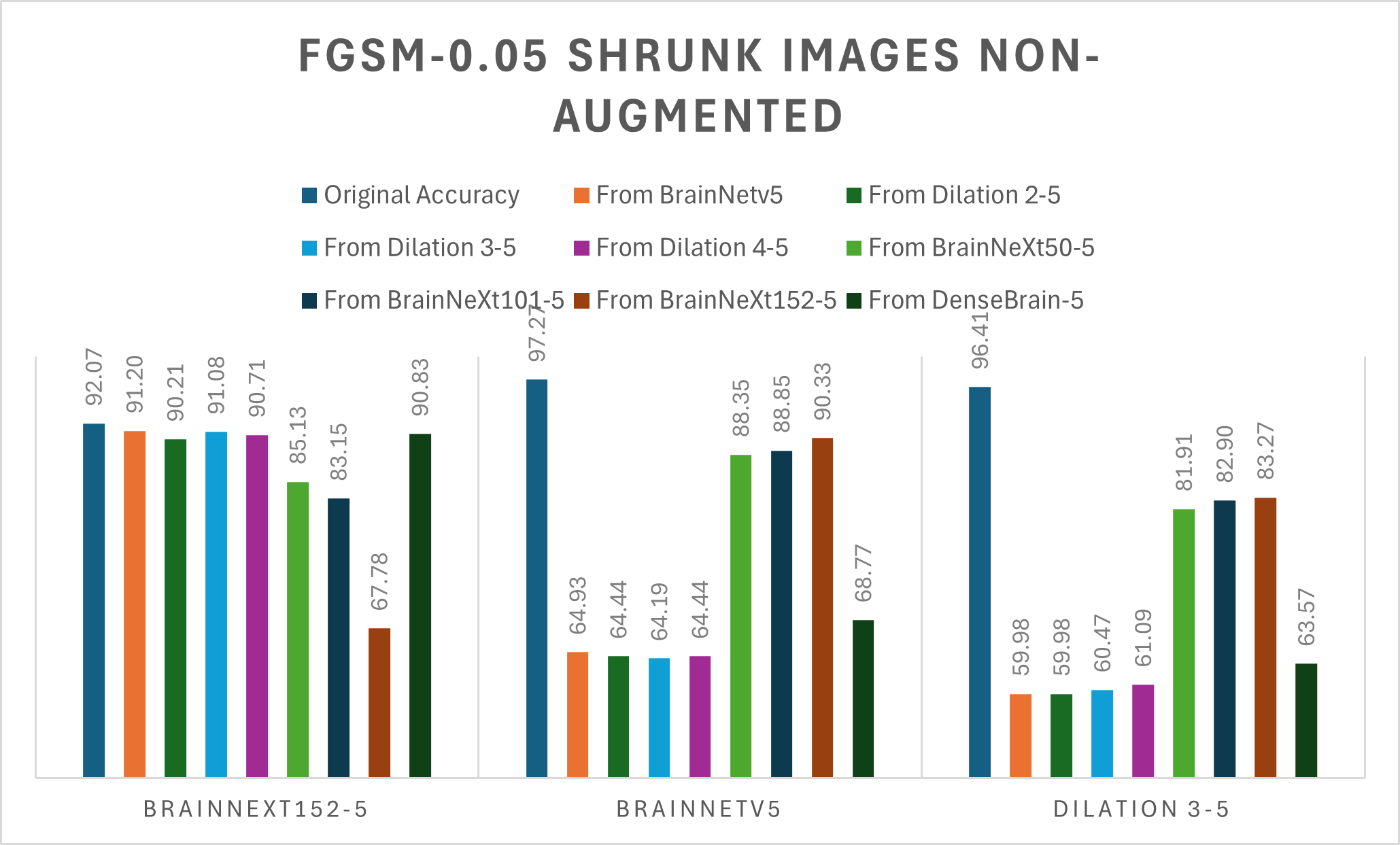}
    \caption{Accuracy of the BrainNeXt152-5, BrainNetv5, and Dilation3-5 models under FGSM white-box and black-box attacks using the shrunk and non-augmented MRI dataset, evaluated at $\epsilon = 0.05$.}
    \label{FGSM-0.05-3}
\end{figure}

\subsection{Results on PGD Attacks}
PGD attacks build upon the strengths of FGSM, leveraging not only the effectiveness of the gradient-based formulation and careful tuning of the $\epsilon$ parameter, but also incorporating a step size $\alpha$ that governs the iterative progression of the attack. This iterative refinement allows PGD to generate stronger, more effective adversarial samples while still controlling for visual imperceptibility. To investigate how the number of iterations and the choice of $\alpha$ influence transferability performance, we conduct a series of experiments using multiple $\alpha$ values, while keeping $\epsilon$ fixed at 0.03 to ensure consistency and attack strength across evaluations. As in the FGSM experiments, we evaluate the robustness of three representative models—BrainNeXt-152, BrainNet, and Dilation3—which exemplify different architectural variants. Adversarial samples are generated from these models and their respective architectural variations (seven models in total), simulating both white-box and black-box scenarios. To assess how input quality and preprocessing affect adversarial effectiveness, we conduct evaluations across the same three dataset variants used in the FGSM setup: (i) full-sized augmented MRI images, (ii) shrunk augmented MRI images (optimized for efficiency) and (iii) shrunk non-augmented MRI images.

\begin{figure}[ht]
   \centering
    \includegraphics[height=0.18\textheight, width=0.45\textwidth]{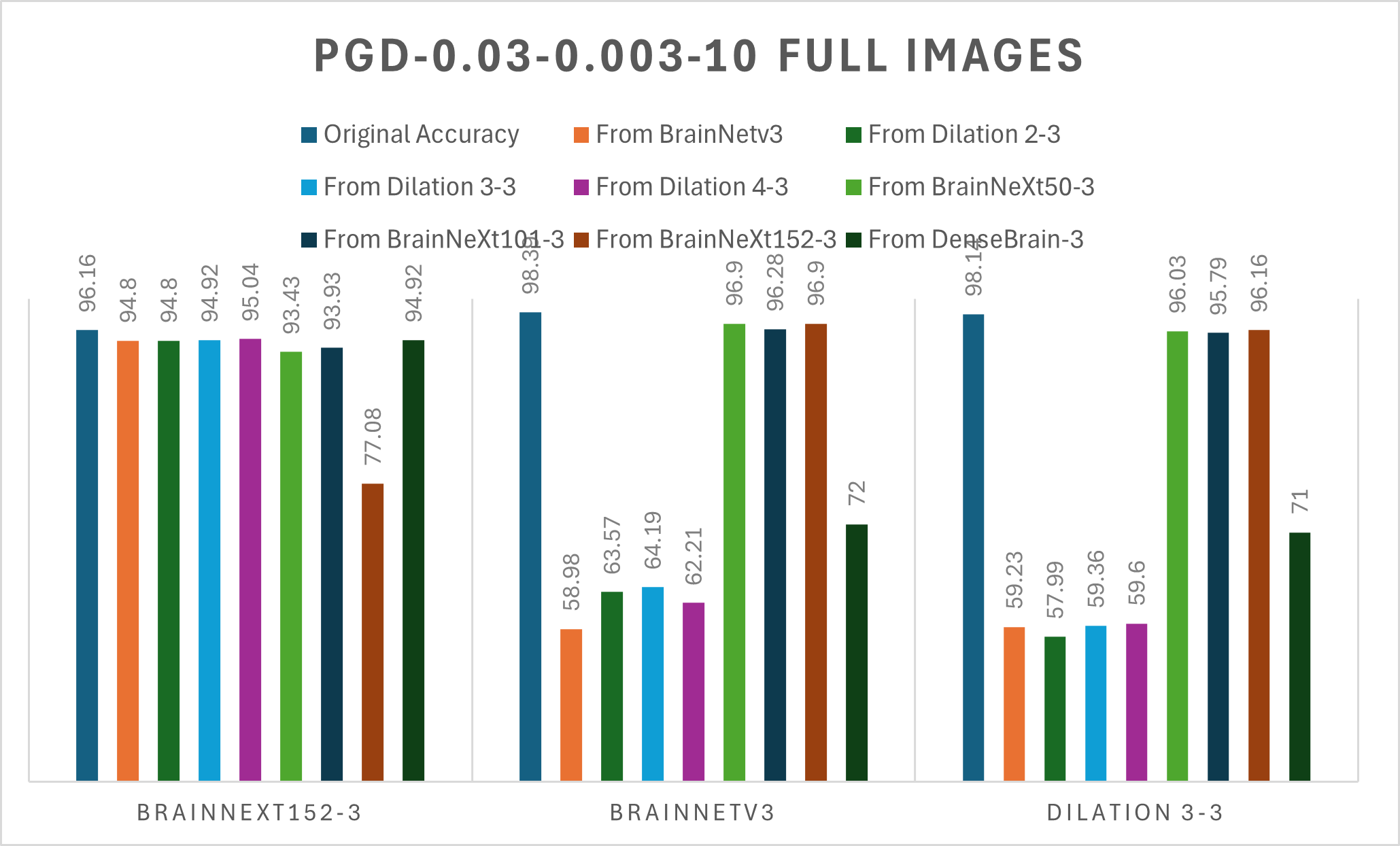}
    \caption{Accuracy of the BrainNeXt152-3, BrainNetv3, and Dilation3-3 models under PGD white-box and black-box attacks using the full-sized and augmented MRI dataset, evaluated at $\epsilon = 0.03$, $\alpha = 0.003$, for 10 iterations.}
    \label{PGD-0.03-0.003-10-1}
\end{figure}
\begin{figure}[ht]
   \centering
    \includegraphics[height=0.18\textheight, width=0.45\textwidth]{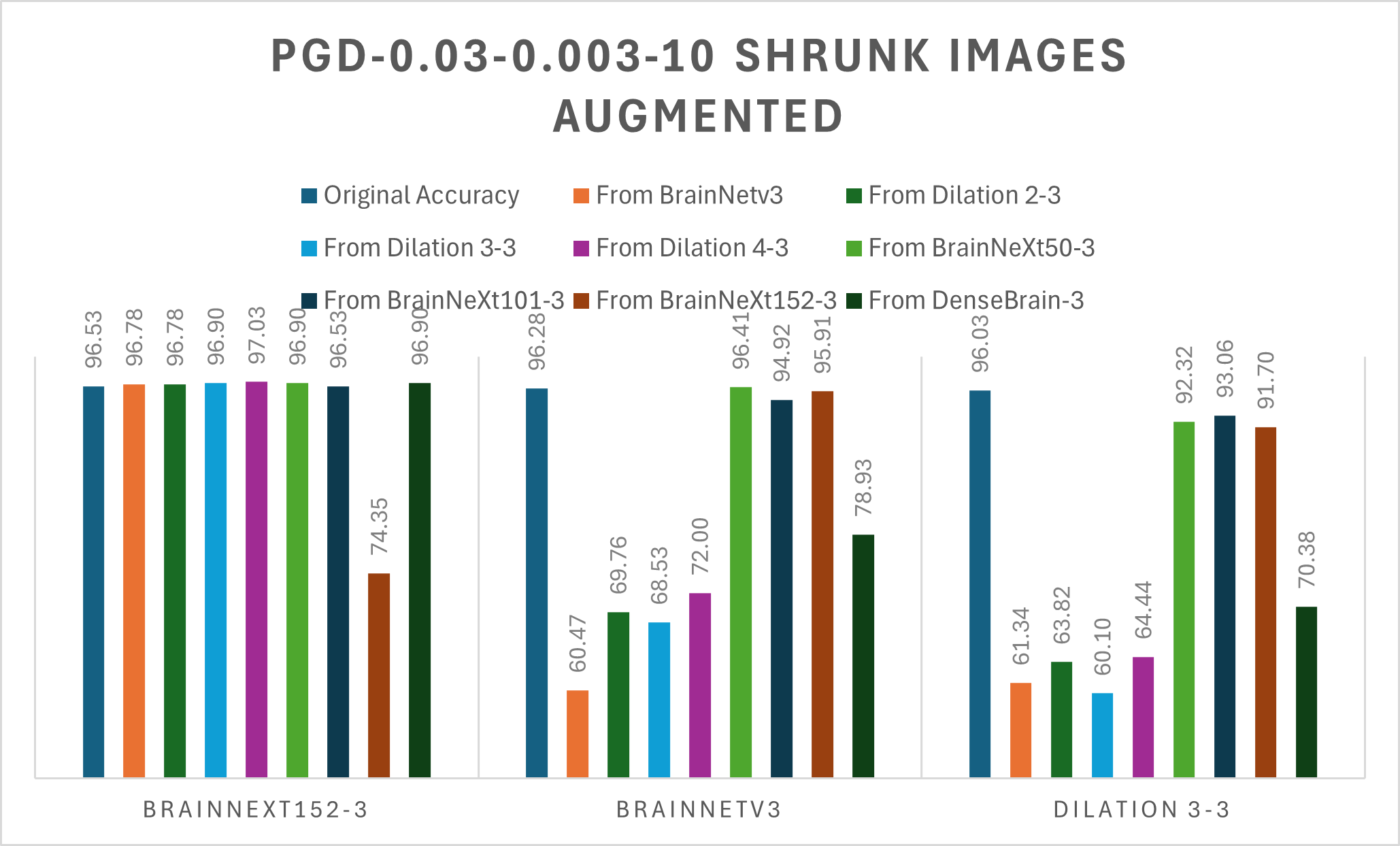}
    \caption{Accuracy of the BrainNeXt152-3, BrainNetv3, and Dilation3-3 models under PGD white-box and black-box attacks using the shrunk and augmented MRI dataset, evaluated at $\epsilon = 0.03$, $\alpha = 0.003$, for 10 iterations.}
    \label{PGD-0.03-0.003-10-2}
\end{figure}
\begin{figure}[ht]
   \centering
    \includegraphics[height=0.18\textheight, width=0.45\textwidth]{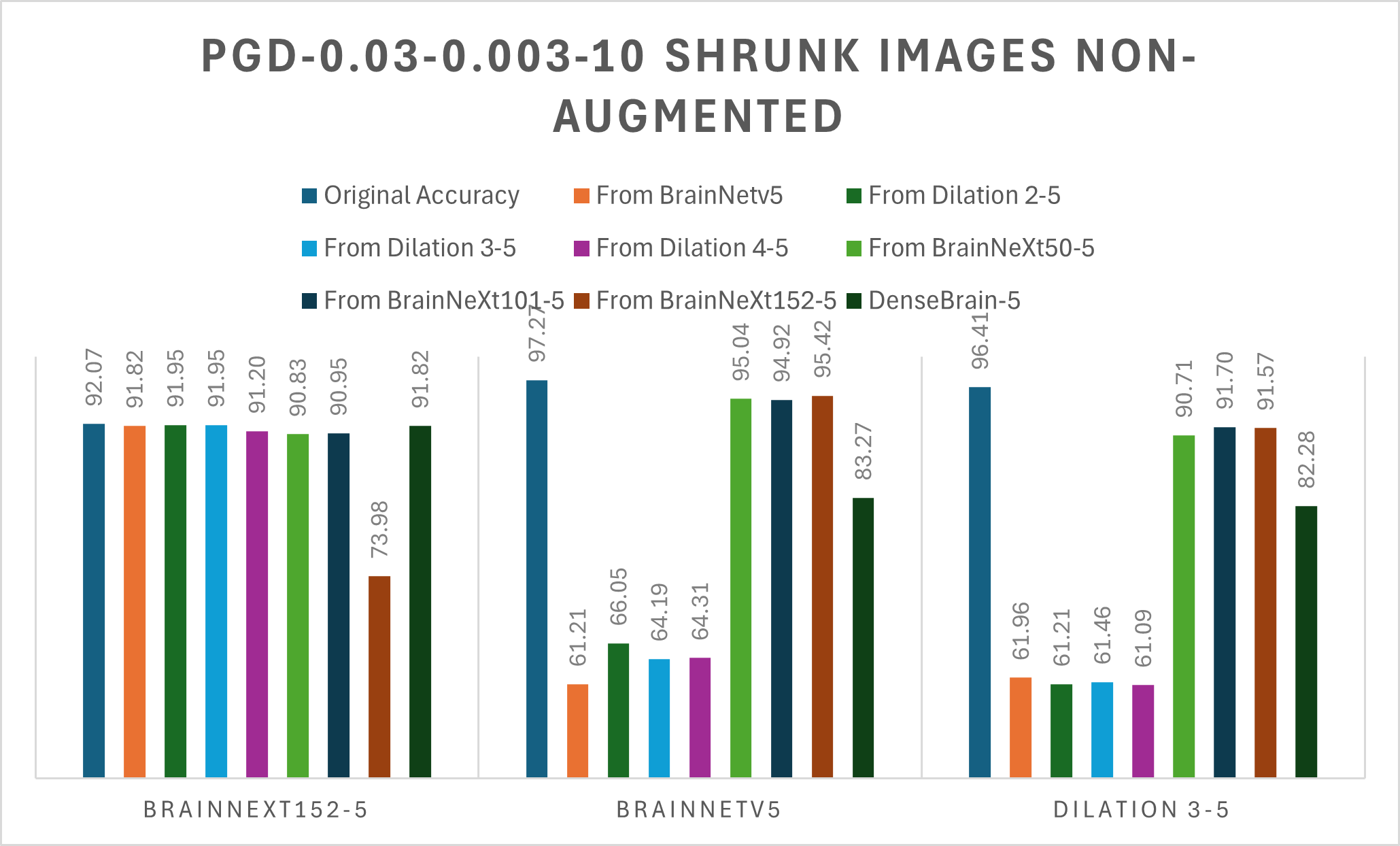}
    \caption{Accuracy of the BrainNeXt152-5, BrainNetv5, and Dilation3-5 models under PGD white-box and black-box attacks using the shrunk and non-augmented MRI dataset, evaluated at $\epsilon = 0.03$, $\alpha = 0.003$, for 10 iterations.}
    \label{PGD-0.03-0.003-10-3}
\end{figure}

Fig.~\ref{PGD-0.03-0.003-10-1}, Fig.~\ref{PGD-0.03-0.003-10-2}, and Fig.~\ref{PGD-0.03-0.003-10-3} illustrate the classification accuracy of the BrainNeXt152-5, BrainNet, and Dilation3 models under PGD-based white-box and black-box attacks, using $\epsilon = 0.03$, $\alpha = 0.003$, and 10 attack iterations. Results are reported across three dataset variants: (i) full-sized augmented, (ii) shrunk augmented, and (iii) shrunk non-augmented MRI images. Across all experiments, BrainNeXt152 once again demonstrates strong resilience against black-box attacks, while simultaneously producing the least transferable adversarial samples, particularly when targeting the BrainNet and Dilation-based models. This pattern is consistent with prior FGSM results, reinforcing the idea that ResNeXt-based architectures, while robust, do not generate highly effective transferable perturbations. In contrast, Dilation3 and BrainNet exhibit more pronounced accuracy drops under black-box attacks generated by other Dilation variants (Dilation2 and Dilation4) and BrainNet itself. These effects are further amplified by PGD’s iterative formulation, where the inclusion of the $\alpha$ parameter refines the attack and enhances transferability while remaining visually imperceptible. The lowest black-box accuracy for Dilation3 is recorded at 61.09\%, while BrainNet's lowest accuracy reaches 64.31\%, both under attacks crafted from the Dilation4 model using the shrunk non-augmented dataset. Overall, excluding attacks from BrainNeXt152, both BrainNet and Dilation3 experience an average drop of approximately 40\% in black-box accuracy on shrunk datasets, with slightly greater degradation on non-augmented images.

Consistent patterns are observed for additional PGD experiments for $\alpha = 0.0075$, tested with 20 iterations. The corresponding results are illustrated in Fig.~\ref{PGD-0.03-0.0075-20-1}, Fig.~\ref{PGD-0.03-0.0075-20-2}, Fig.~\ref{PGD-0.03-0.0075-20-3}, respectively. These figures show the classification accuracy of the BrainNeXt152-5, BrainNet, and Dilation3 models under PGD white-box and black-box attacks, evaluated across the same three dataset configurations as in previous experiments. Notably, for $\alpha = 0.0075$, we observe the largest drops in accuracy for both Dilation3 and BrainNet under black-box attacks. Specifically, Dilation3 drops to 60.35\% when attacked with adversarial samples generated by Dilation4, while BrainNet reaches its lowest accuracy of 58.36\% when targeted by samples from Dilation3—both under the shrunk non-augmented MRI data setting. These results not only reinforce our earlier finding from the FGSM experiments (i.e., that shrunk non-augmented data leads to reduced robustness against black-box attacks), but also emphasize the impact of a higher $\alpha$ value combined with more iterations in crafting stronger and more transferable adversarial attacks. Importantly, both Dilation3 and BrainNet continue to show strong resilience against attacks generated by BrainNeXt variants, with their accuracies remaining above 88\% in all such cases. Lastly, it must be noted that across all PGD experiments, BrainNeXt demonstrated strong resilience against DenseNet-based attacks, while both Dilation3 and BrainNet suffered significant performance degradation, consistent with the trends observed in the FGSM experiments.


\begin{figure}[ht]
   \centering
    \includegraphics[height=0.18\textheight, width=0.45\textwidth]{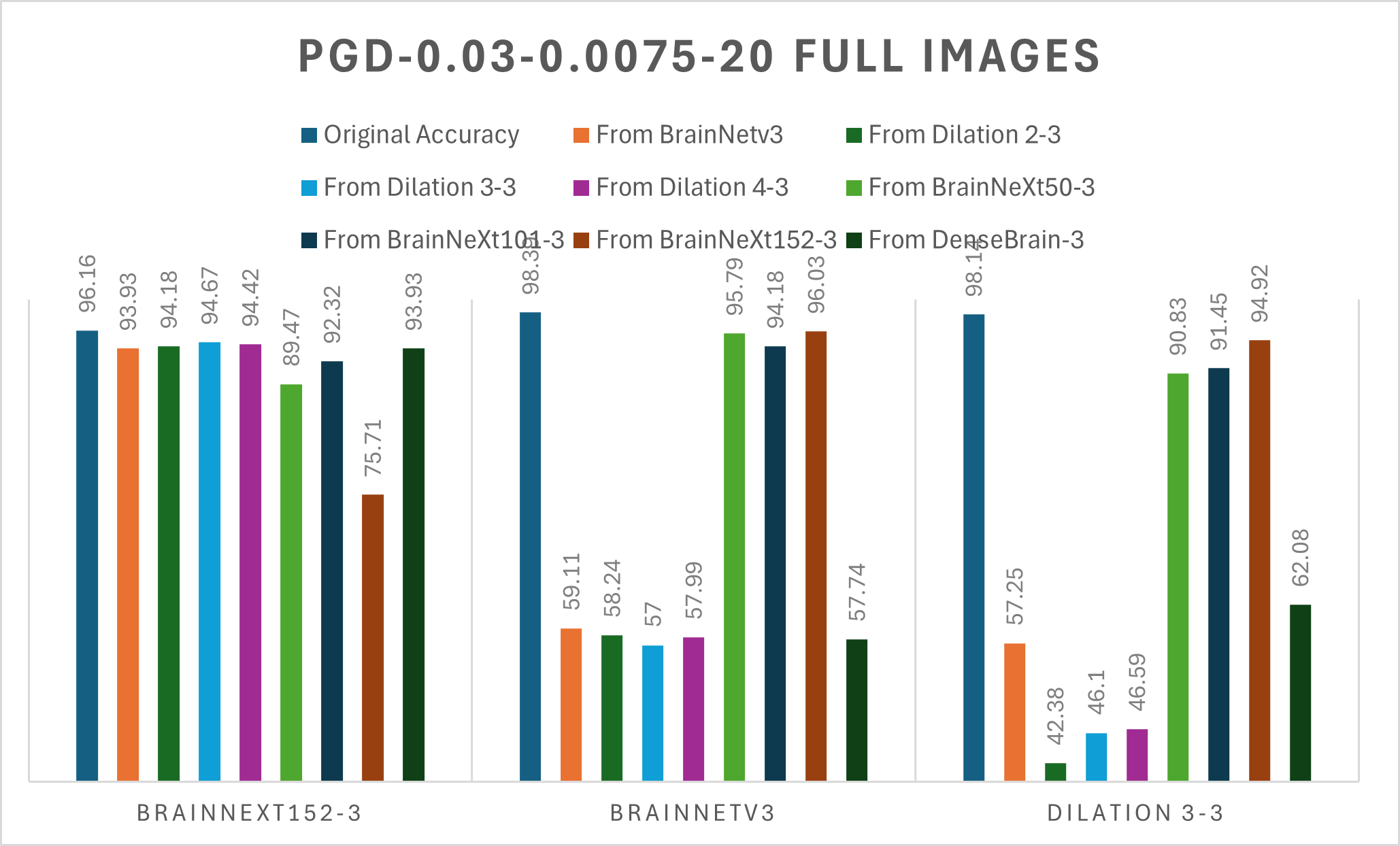}
    \caption{Accuracy of the BrainNeXt152-3, BrainNetv3, and Dilation3-3 models under PGD white-box and black-box attacks using the full-sized and augmented MRI dataset, evaluated at $\epsilon = 0.03$, $\alpha = 0.0075$, for 20 iterations.}
    \label{PGD-0.03-0.0075-20-1}
\end{figure}
\begin{figure}[ht]
   \centering
    \includegraphics[height=0.18\textheight, width=0.45\textwidth]{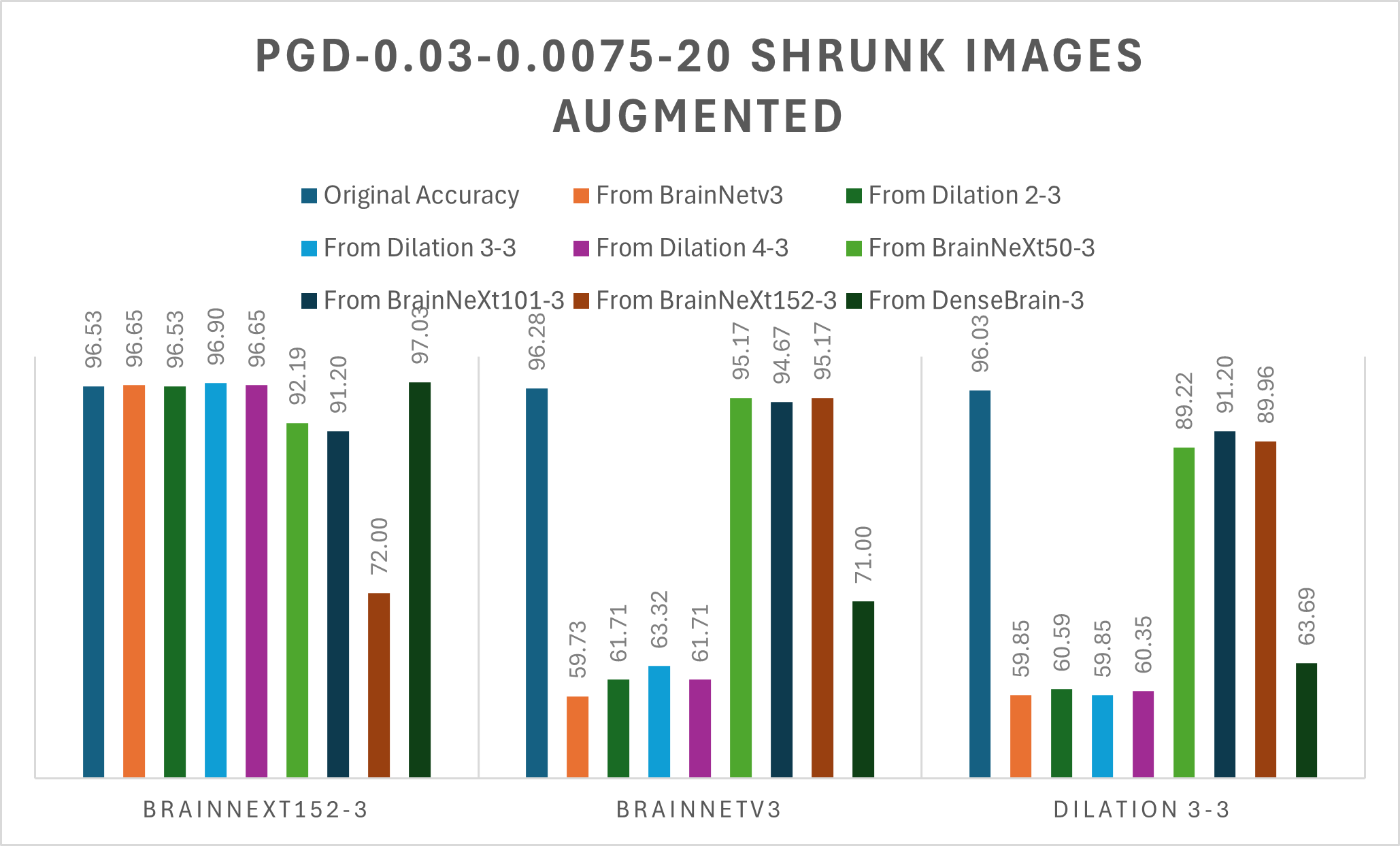}
    \caption{Accuracy of the BrainNeXt152-3, BrainNetv3, and Dilation3-3 models under PGD white-box and black-box attacks using the shrunk and augmented MRI dataset, evaluated at $\epsilon = 0.03$, $\alpha = 0.0075$, for 20 iterations.}
    \label{PGD-0.03-0.0075-20-2}
\end{figure}
\begin{figure}[ht]
   \centering
    \includegraphics[height=0.18\textheight, width=0.45\textwidth]{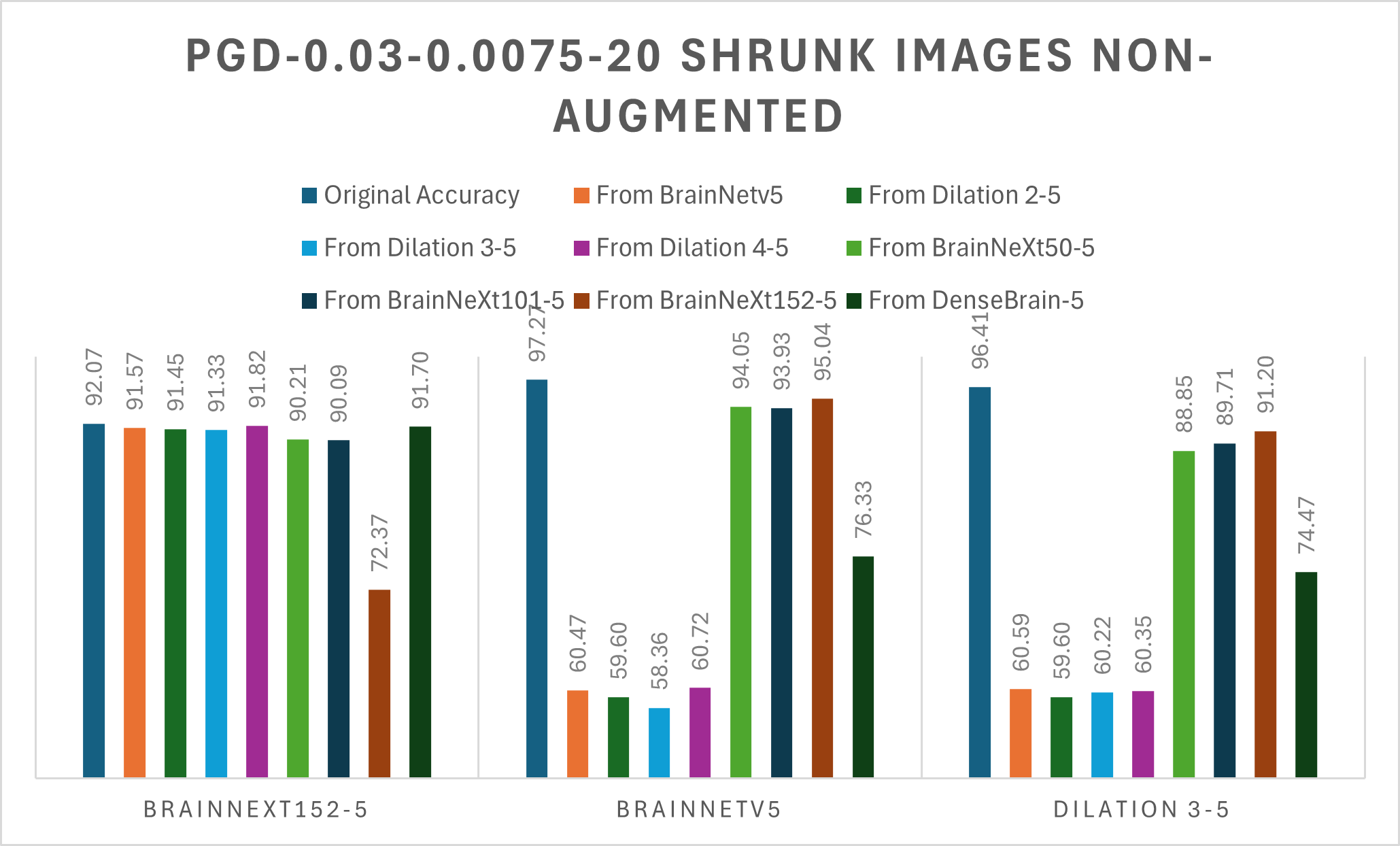}
    \caption{Accuracy of the BrainNeXt152-5, BrainNetv5, and Dilation3-5 models under PGD white-box and black-box attacks using the shrunk and non-augmented MRI dataset, evaluated at $\epsilon = 0.03$, $\alpha = 0.0075$, for 20 iterations.}
    \label{PGD-0.03-0.0075-20-3}
\end{figure}

\subsection{Analysis of Results \& Discussion}
The experimental results revealed several consistent trends. First, models based on the ResNeXt architecture (e.g., BrainNeXt152) exhibited stronger resilience to black-box adversarial attacks originating from DenseNet, ResNet and Dilation-based models, as well as a degree of self-resilience. However, these models also generated comparatively weaker adversarial examples when attacking other architectures. This maybe attributed to ResNeXt’s use of cardinality, i.e., multiple parallel transformation paths via grouped convolutions, which encourages richer and more diverse feature representations compared to standard ResNet architectures. Additionally, the smaller filter sizes within these paths may help the model focus on fine-grained spatial patterns, which is an advantage particularly relevant in MRI-based brain tumor classification. In contrast, ResNet and DilationResNet models produced stronger and more transferable attacks against each other, likely due to their architectural similarities and shared feature extraction pathways. Similarly, both models exhibited weaknesses against both FGSM and PGD attacks generated from the DenseNet model, with accuracy drops ranging from approximately 22\% to 30\%.

Furthermore, we observed that increasing the perturbation magnitude via higher $\epsilon$ in FGSM or higher $\alpha$ in PGD, led to lower classification accuracy across all models. This aligns with expectations, as stronger perturbations push inputs beyond decision boundaries more effectively. Notably, the most significant declines in robustness occurred when models were trained on shrunk, non-augmented MRI images. While these models maintained high accuracy on untampered test data, they proved highly vulnerable to adversarial noise. This suggests that reduced image resolution and lack of augmentation stripped the model of exposure to spatial and contextual variability, which is especially detrimental in MRI-based tumor classification tasks where subtle spatial features are critical for robust learning.
\section{Conclusions \& Future Work}
\label{s:conc}

In this study, we investigated the adversarial robustness of three deep learning model families, namely BrainNet (ResNet-based), BrainNeXt (ResNeXt-based) and DilationNet (dilated convolution variants) for MRI-based brain tumor classification. While BrainNeXt models exhibited the highest resilience under black-box adversarial attacks, particularly across varied datasets and attack parameters, they also produced the least transferable adversarial samples. Conversely, BrainNet and Dilation models demonstrated higher vulnerability to transferable attacks, especially in low-resolution, non-augmented settings, despite maintaining strong baseline accuracy. These results underscore how both architectural design and data quality (e.g., resolution and augmentation) critically influence model robustness in clinical imaging tasks. Future work will focus on expanding our evaluation to include more adaptive and targeted attack strategies \cite{aina-2025}, as well as exploring defense mechanisms such as adversarial training and attention-guided regularization. We also plan to investigate robustness-aware neural architecture search methods, aiming to identify architectures that are not only accurate and efficient, but inherently more resilient under real-world black-box settings in the medical domain.

\section{Appendix A - Additional Results on FGSM}
In this appendix section, we firstly attach the results from additional experiments we conducted, using $\epsilon=0.02$ as the perturbation parameter for these attacks. The accuracies achieved by each model after the attacks were applied follow the same pattern that was observed in the main section. In particular, once again the models that were based on the ResNeXt architecture exhibit stronger resilience to black-box adversarial attacks originating from other architectures. In addition, the attacks originating from ResNeXt-based models are once again less effective against other architectures.

\begin{figure}[ht]
   \centering
    \includegraphics[height=0.2\textheight, width=0.45\textwidth]{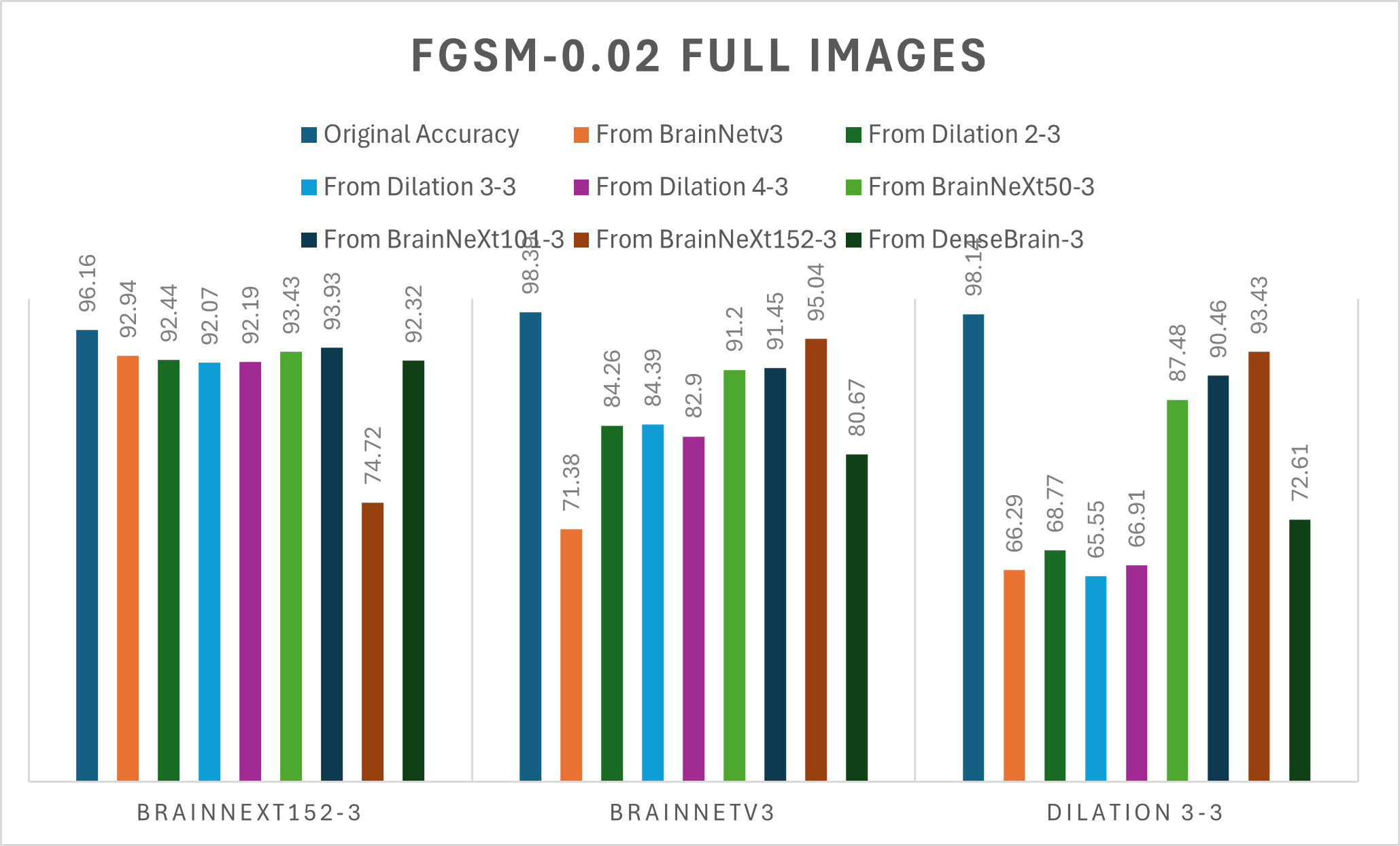}
    \caption{Accuracy of the BrainNeXt152-5, BrainNetv5, and Dilation3-5 models under FGSM white-box and black-box attacks using the full-sized and augmented MRI dataset, evaluated at $\epsilon = 0.02$.}
    \label{FGSM-0.02-1}
\end{figure}
\begin{figure}[ht]
   \centering
    \includegraphics[height=0.2\textheight, width=0.45\textwidth]{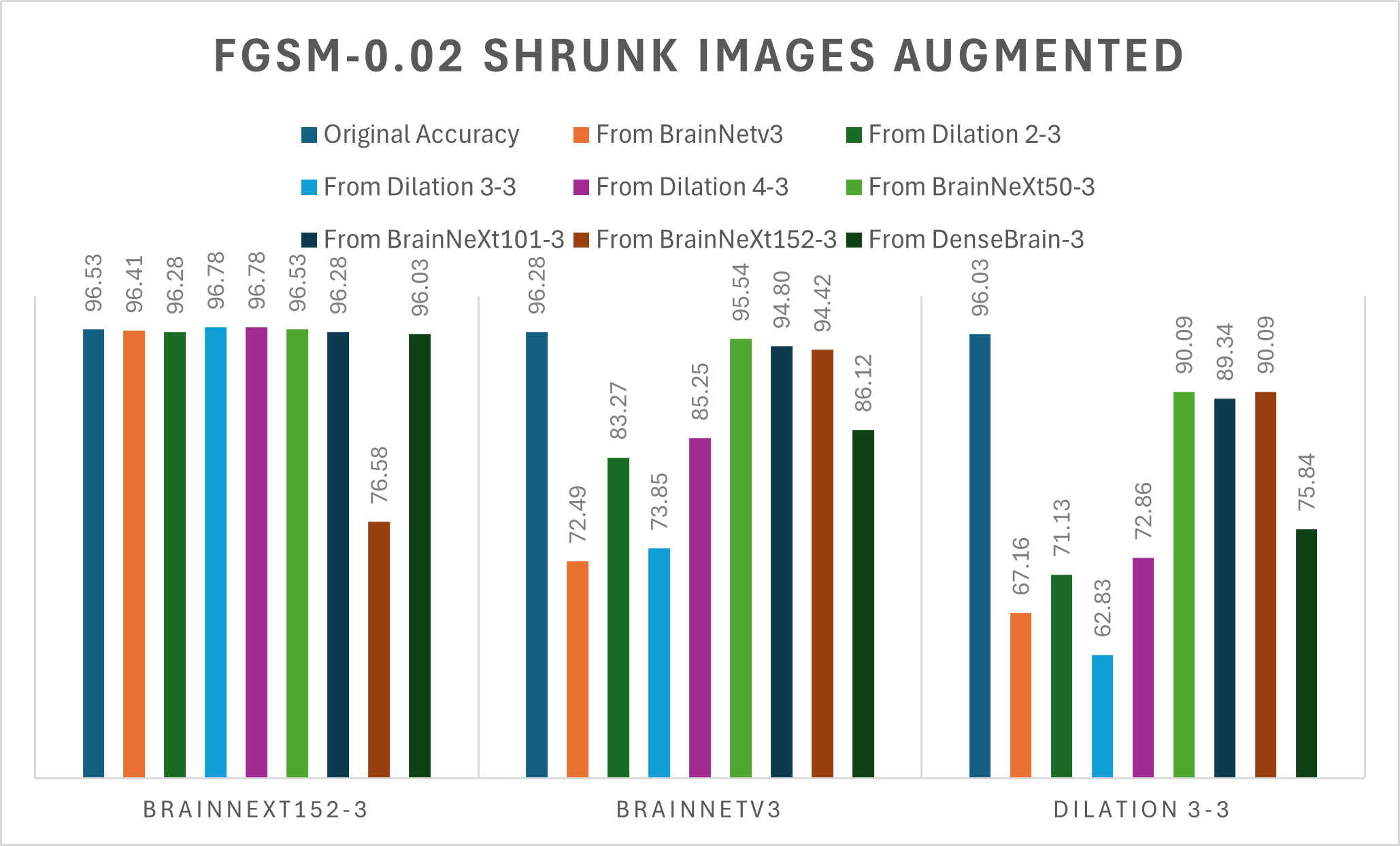}
    \caption{Accuracy of the BrainNeXt152-5, BrainNetv5, and Dilation3-5 models under FGSM white-box and black-box attacks using the shrunk and augmented MRI dataset, evaluated at $\epsilon = 0.02$.}
    \label{FGSM-0.02-2}
\end{figure}
\begin{figure}[ht]
   \centering
    \includegraphics[height=0.2\textheight, width=0.45\textwidth]{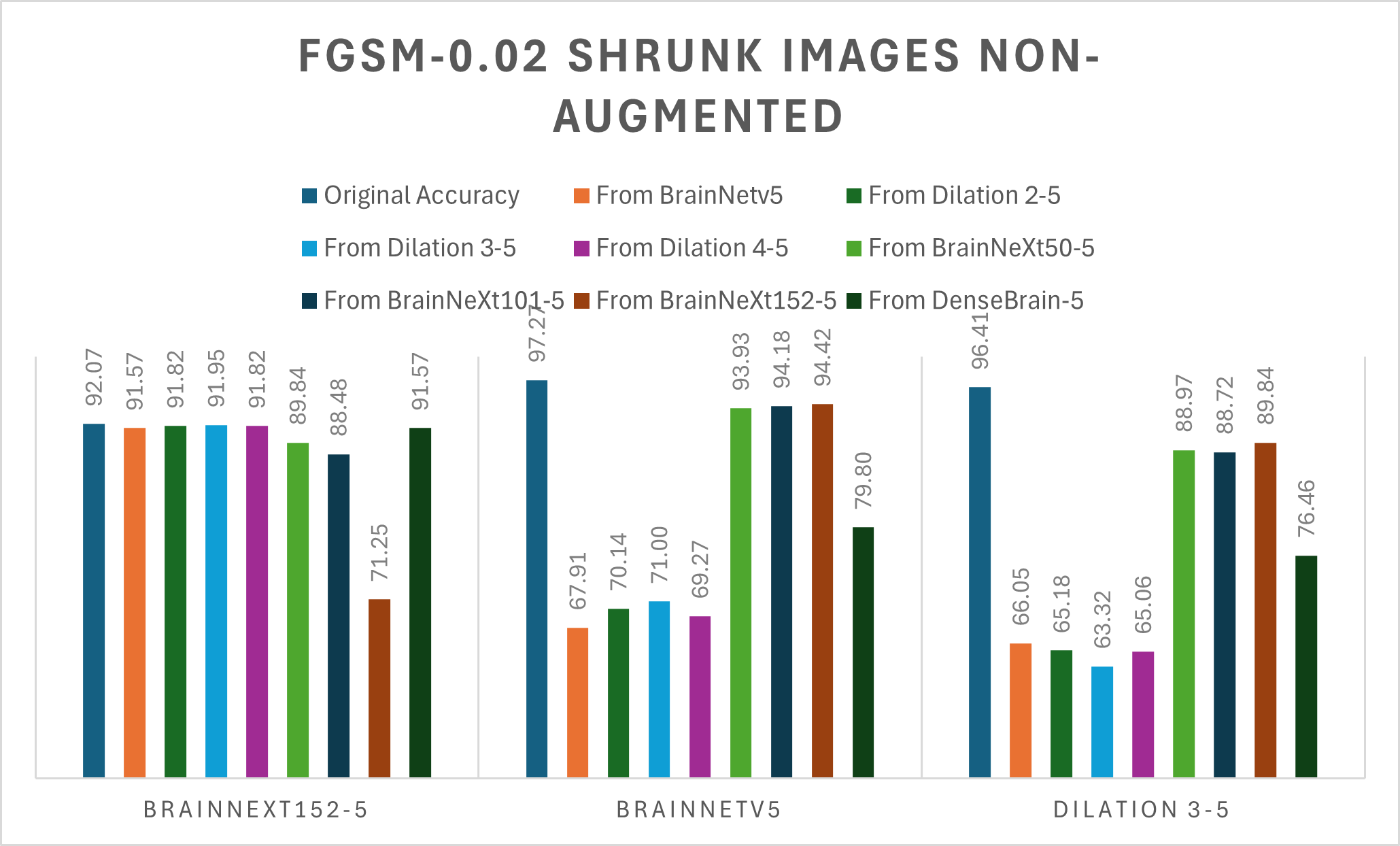}
    \caption{Accuracy of the BrainNeXt152-5, BrainNetv5, and Dilation3-5 models under FGSM white-box and black-box attacks using the shrunk and non-augmented MRI dataset, evaluated at $\epsilon = 0.02$.}
    \label{FGSM-0.02-3}
\end{figure}

\section{Appendix B - Additional Results on PGD}
A similar pattern is also observed in our additional experiments when using the PGD attack for $\alpha=0.0015$ for 20 iterations, with ResNeXt-based architectures exhibiting stronger resilience to black-box adversarial attacks originating from other architectures, but weaker transferability of their own. 

\begin{figure}[ht]
   \centering
    \includegraphics[height=0.2\textheight, width=0.45\textwidth]{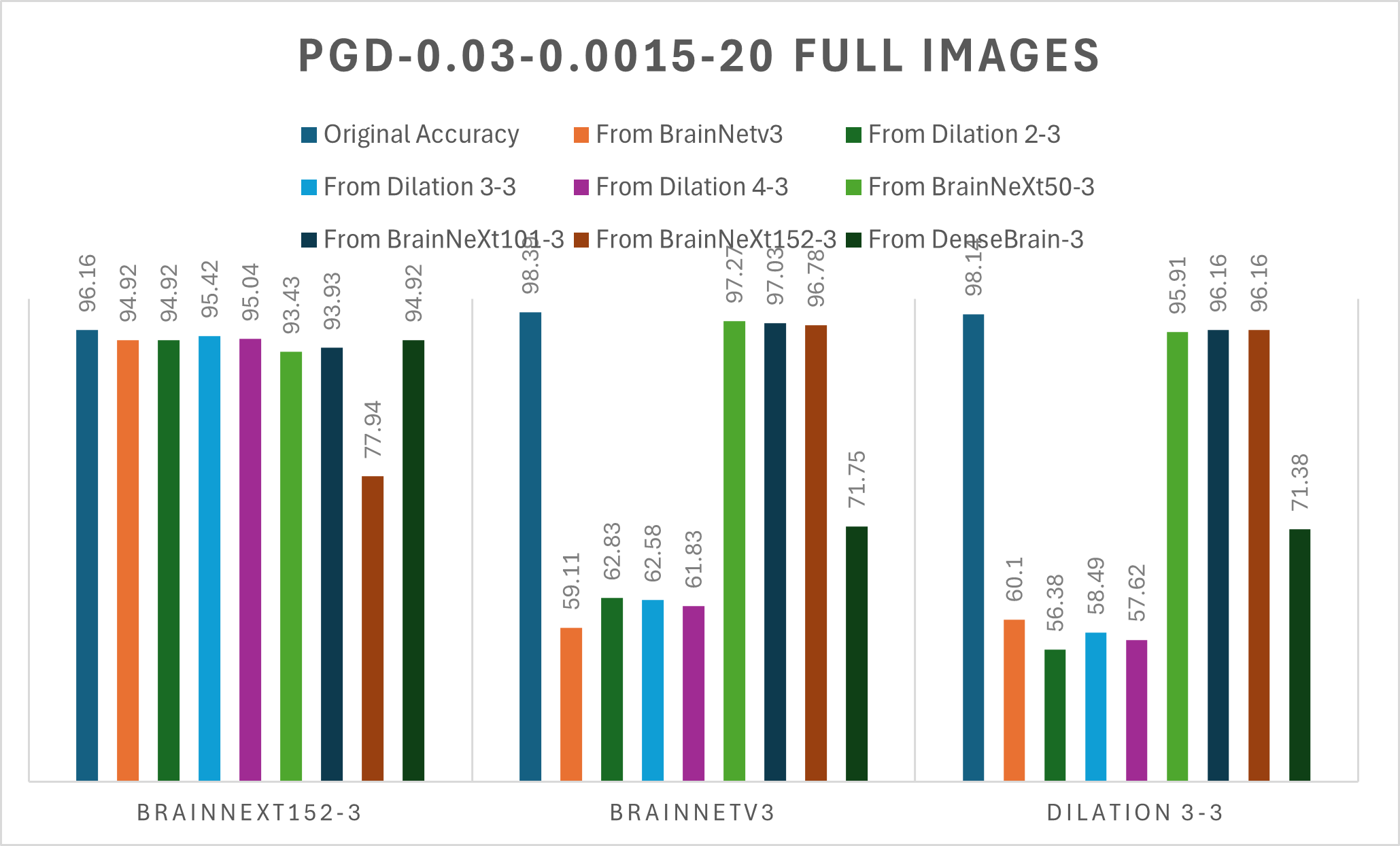}
    \caption{Accuracy of the BrainNeXt152-5, BrainNetv5, and Dilation3-5 models under PGD white-box and black-box attacks using the full-sized and augmented MRI dataset, evaluated at $\epsilon = 0.03$, $\alpha = 0.0015$, for 20 iterations.}
    \label{PGD-0.03-0.0015-20-1}
\end{figure}
\begin{figure}[ht]
   \centering
    \includegraphics[height=0.2\textheight, width=0.45\textwidth]{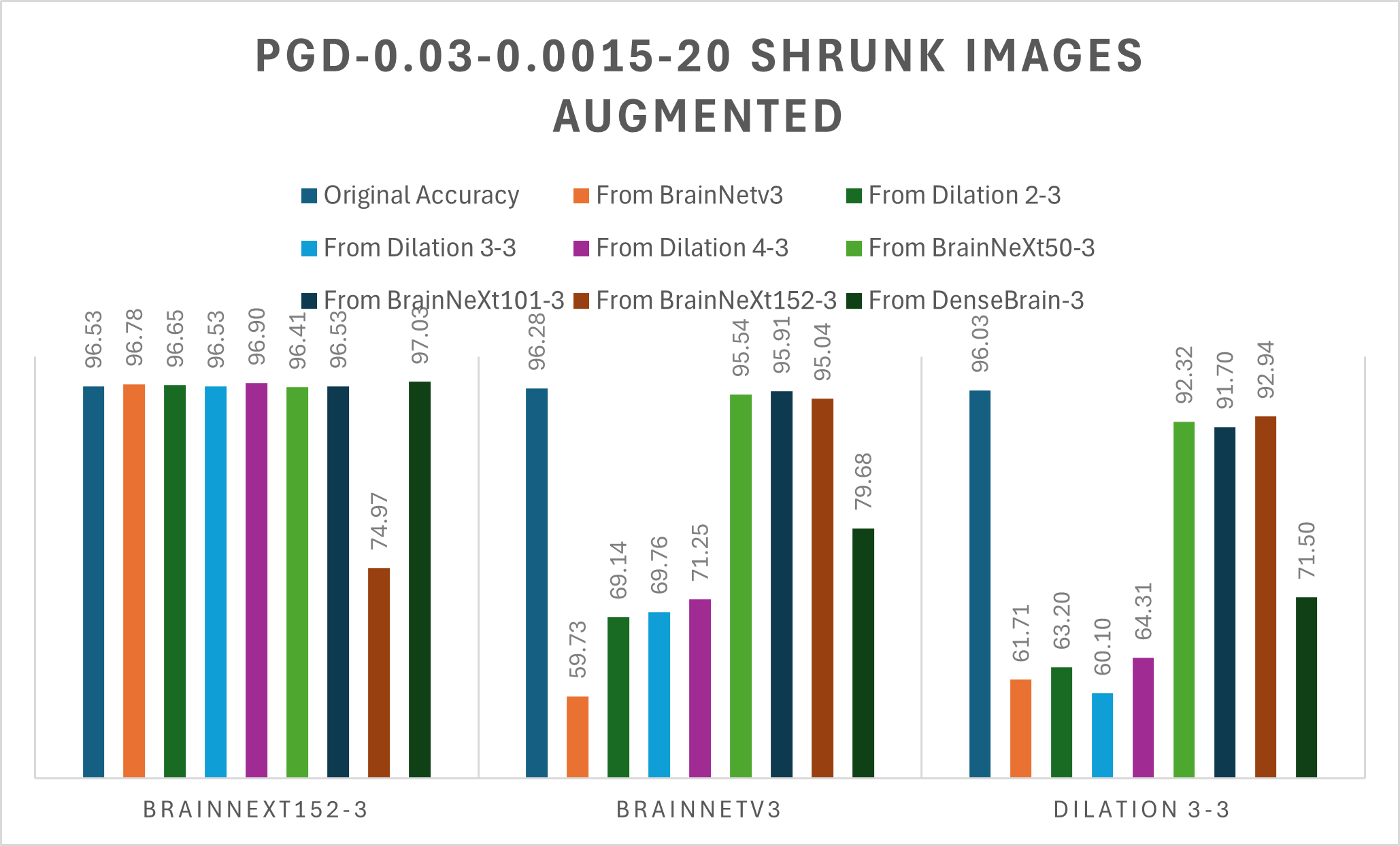}
    \caption{Accuracy of the BrainNeXt152-5, BrainNetv5, and Dilation3-5 models under PGD white-box and black-box attacks using the shrunk and augmented MRI dataset, evaluated at $\epsilon = 0.03$, $\alpha = 0.0015$, for 20 iterations.}
    \label{PGD-0.03-0.0015-20-2}
\end{figure}
\begin{figure}[ht]
   \centering
    \includegraphics[height=0.2\textheight, width=0.45\textwidth]{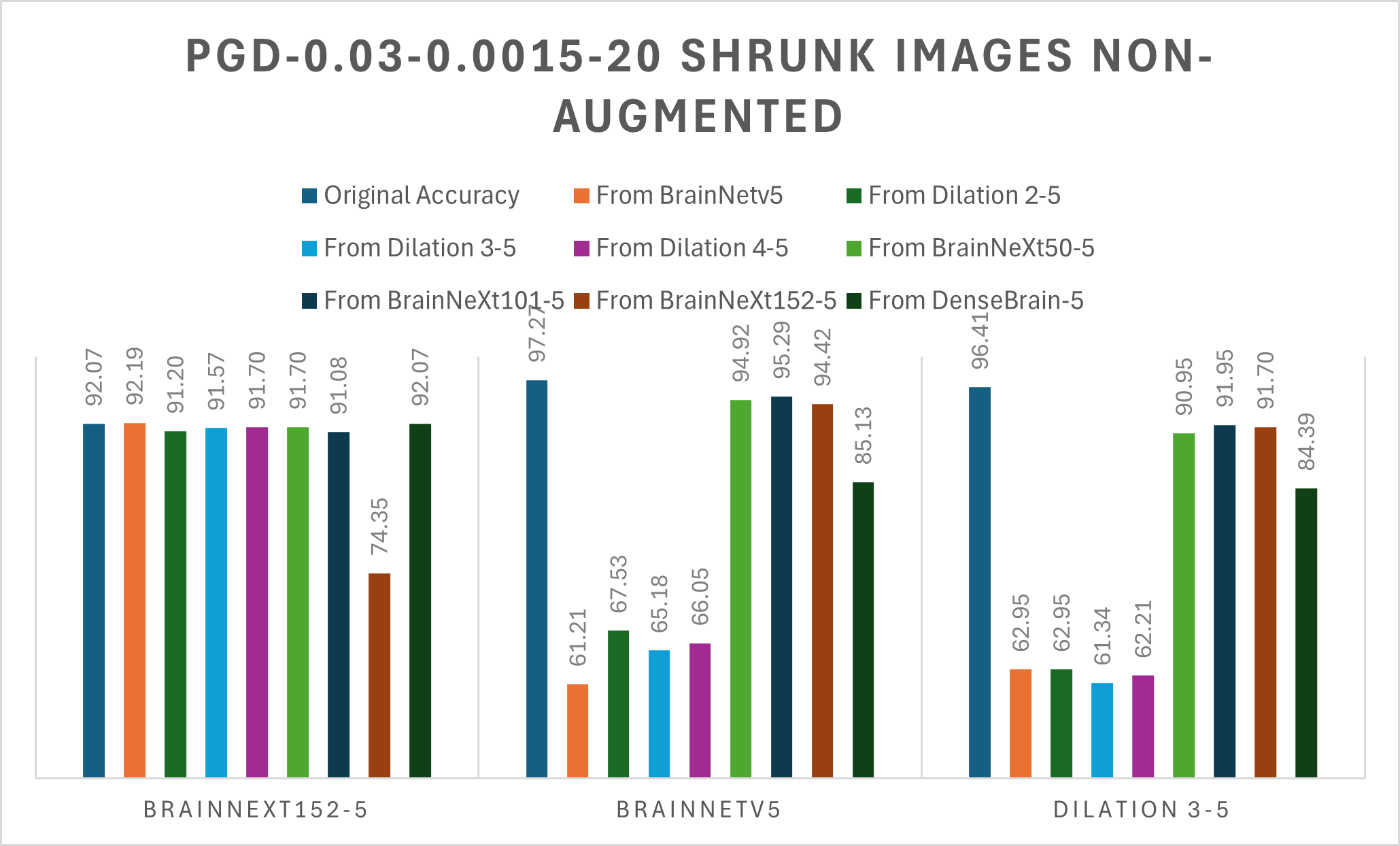}
    \caption{Accuracy of the BrainNeXt152-5, BrainNetv5, and Dilation3-5 models under PGD white-box and black-box attacks using the shrunk and non-augmented MRI dataset, evaluated at $\epsilon = 0.03$, $\alpha = 0.0015$, for 20 iterations.}
    \label{PGD-0.03-0.0015-20-3}
\end{figure}




\nocite{}
\bibliographystyle{IEEEtran}
\bibliography{references}
\end{document}